\documentclass{elsarticle}
\usepackage[caption=false]{subfig}
\usepackage{lipsum}
\usepackage{amsmath,amssymb,amsfonts}
\usepackage{graphicx}
\usepackage{epstopdf}
\usepackage{algorithmic}
\usepackage{algorithm}
\usepackage{url}
\usepackage{array}

\usepackage{lineno,hyperref}
\modulolinenumbers[5]

\journal{Pattern Recognition}

\begin{document}

\begin{frontmatter}

\title{Detection of moving objects through turbulent media. Decomposition of Oscillatory vs Non-Oscillatory spatio-temporal vector fields}

\author[sdsu]{J. Gilles\corref{mycorrespondingauthor}}
\ead{jgilles@sdsu.edu}
\ead[url]{http://jegilles.sdsu.edu}

\author[sdsu]{F. Alvarez}
\ead{aalvarez.fr@gmail.com}

\author[sdsu]{N. Ferrante}
\ead{nicholas.b.ferrante@icloud.com}

\author[lakeforest]{M. Fortman}
\ead{fortmanma@mx.lakeforest.edu}

\author[millscollege]{L. Tahir}
\ead{ltahir@mills.edu}

\author[creighton]{A. Tarter}
\ead{act72583@creighton.edu}

\author[georgetown]{A. von Seeger}
\ead{amv48@georgetown.edu}

\address[sdsu]{Department of Mathematics \& Statistics, San Diego State University, San Diego, 5500 Campanile Dr, CA 92182, USA}
\address[lakeforest]{Lake Forest College, LFC \#582, 555 N Sheridan Rd, Lake Forest, IL 60045, USA}
\address[millscollege]{Mills College, 5000 MacArthur Blvd, Oakland, CA 94613, USA}
\address[creighton]{Creighton University, 2500 California Plaza, Omaha, NE 68178, USA}
\address[georgetown]{Department of Mathematics, Georgetown University, 327A St. Mary's Hall, 3700 O St NW, Washington D.C. 20057, USA}

\cortext[mycorrespondingauthor]{Corresponding author}

\begin{abstract}
In this paper, we investigate how moving objects can be detected when images are impacted by atmospheric turbulence. We present a geometric spatio-temporal point of view to the 
problem and show that it is possible to distinguish movement due to the turbulence vs. moving objects. To perform this task, we propose an extension of 2D cartoon+texture 
decomposition algorithms to 3D vector fields. Our algorithm is based on curvelet spaces which permit to better characterize the movement flow geometry. We present experiments on real 
data which illustrate the efficiency of the proposed method.
\end{abstract}


\end{frontmatter}

\linenumbers

\section{Introduction}
Detection of moving objects in sequences of images is a standard problem in video processing, notably for video surveillance or military applications. Generally such algorithms are 
based on two main steps: the first step detects all moving objects while the second step performs the tracking of these different objects and permits to reject objects with erratic movements. The second step is usually based on some 
adaptive filtering (e.g. Kalman filter \cite{kalman}) and will not be investigated in this paper. The algorithms used to perform the first step can be divided into two main 
families: background subtraction or optical flow based methods. Background subtraction methods aim at estimating the fixed background in the sequence \cite{bck2,bck1}. The 
estimated background is then subtracted from each frame and the resulting sequence is then thresholded, providing only clusters of pixels corresponding to changes, i.e moving 
objects. Optical flow based techniques \cite{Black1996,Horn1981,DiffeoDemons,ipolHS,Zach,ipolTVL1} aim at estimating the velocity vector field corresponding to movement present in the 
sequence, i.e. each pixel contains a vector representing the displacement of that pixel from the current frame to the next one.
In this paper, we consider the detection of moving objects through observations affected by atmospheric turbulence. It is straightforward to imagine that the presence of turbulence will create additional movement over the whole image, hence challenging the previously cited methods.\\
Imaging through atmospheric turbulence has generated many contributions, mainly for resto\-ration purposes 
\cite{Anantrasirichai2013,Frakes2001,Gepshtein2004,Gilles2012,waveburst,Halder,Halder2,Halder3,Li2007,Lou2013,MaoGilles,Mario3,Mario1,Song,Yang,Zhu2010,Zwart} just to cite a few. 
The addressed question was how to get a ``clean'' image from a sequence of observations which are distorted by the atmospheric turbulence. Interestingly, the problem of detecting 
moving objects through atmospheric turbulence has been investigated only by a few people. Most contributions introduce modifications of the background subtraction method and more importantly they design very specific tracking steps to take into account the outliers introduced by the turbulence. For instance, in 
\cite{Chen2014}, the authors propose to use an adaptive thresholding technique to distinguish clusters of turbulence movement vs. object movement. The same type of approach is 
also used in \cite{Apuroop,Robinson2016}. Experimentally, we can observe 
that this idea does not work well when the velocity magnitudes of both types of movement are of the same level. A two-level thresholding technique combined with a neural network is 
described in \cite{halder2016moving}; the challenging aspect to this method is how to train the neural network to be efficient in most situations. In \cite{Oreifej}, the authors 
extend the background subtraction technique to a background+moving objects+turbulence movement separation model by using a low-rank+sparse decomposition approach. This method is 
directly applied on the sequence frames and does not use any velocity information. Moreover, it requires an a priori model to make the distinction between moving objects and turbulence but 
such a model is not necessarily easy to build. It is interesting to notice the work in \cite{fishbain} where the authors' initial purpose is to perform super-resolution while 
taking into account moving objects. The proposed algorithm includes a step which aims at identifying moving objects. To do so, the authors propose to compute the standard 
deviation of the velocity vector orientations within a small neighborhood. The key idea is that the moving object must have consistent velocity orientations, whereas the turbulence 
velocity is more chaotic. \\
In this paper, we propose a new approach, also based on the idea of consistency, to separate movement due to the turbulence vs. moving objects. Our key idea is to 
consider the 2D+Time movement vector field as a whole and notice that a consistent movement corresponds to a geometric structure while the turbulence movement corresponds 
to oscillating vectors. We then propose to decompose the original vector flow into a geometric vector flow and an oscillating vector flow. Techniques decomposing non-oscillating 
v.s oscillating component were widely studied in image processing to decompose images into their cartoon and textures parts \cite{meyer}. The main idea is to decompose an image 
$f$ into its geometric and textures components, $u$ and $v$, respectively, such that $f=u+v$. Such decomposition is performed by minimizing a functional like 
$$(\hat{u},\hat{v})=\arg_{u\in X,v\in Y}\min \|u\|_X+\lambda\|v\|_Y,$$
where the spaces $X$ and $Y$ are well chosen to correspond to the expected characteristis of each components. More details will be given in Section~\ref{sec:dec}. Some vector flow 
decomposition 
methods were proposed in \cite{Steidl05,Steidl07}. The authors aim to decompose flows into their divergence and curl free components for the purpose of extracting fluid motion information useful in fluid mechanics. In this paper, we propose a different 
type of decomposition which focuses on separating the non-oscillating and oscillating component of a vector field. To perform such task, our model minimizes a functional based on curvelet spaces norms. 
This permits to better characterize the presence of geometric structures.\\
The rest 
of the paper is organized as follows. Section~\ref{sec:media} introduces a geometrical point of view of consistency and explains why this problem can be solved as a vector field 
decomposition problem. Section~\ref{sec:dec} recalls decomposition methods in image processing and introduces our decomposition model for spatio-temporal vector fields. In Section~\ref{sec:exp}, we will experimentally investigate the decomposition efficiency, the impact of the chosen movement flow estimation technique, the impact on the detected object, as well as comparison with a background subtraction technique. Finally, Section~\ref{sec:conc} will conclude the proposed work.

\section{Moving objects in turbulent media}\label{sec:media}
\begin{figure}[!t]
\centering
\begin{tabular}{m{0.15\textwidth}m{0.3\textwidth}m{0.3\textwidth}m{0.1\textwidth}}
Frame~10 & \includegraphics[width=0.3\textwidth]{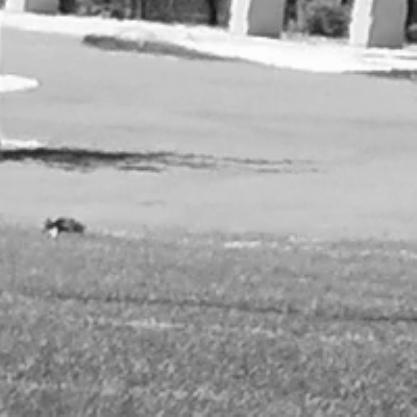} & \includegraphics[width=0.3\textwidth]{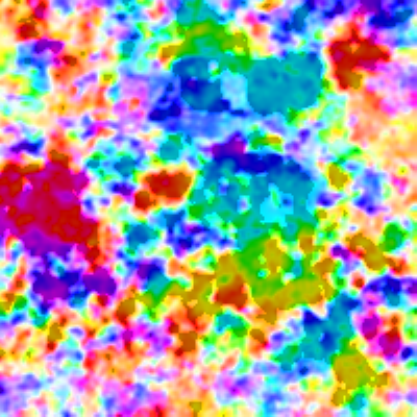} & \\
Frame~30 & \includegraphics[width=0.3\textwidth]{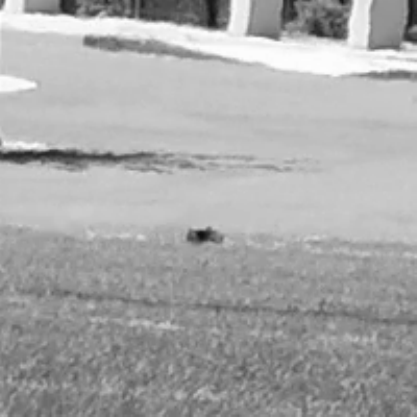} & \includegraphics[width=0.3\textwidth]{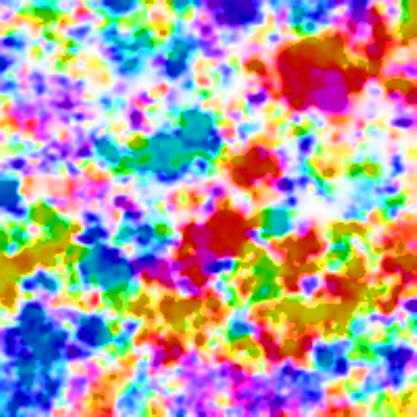} & \includegraphics[scale=0.07]{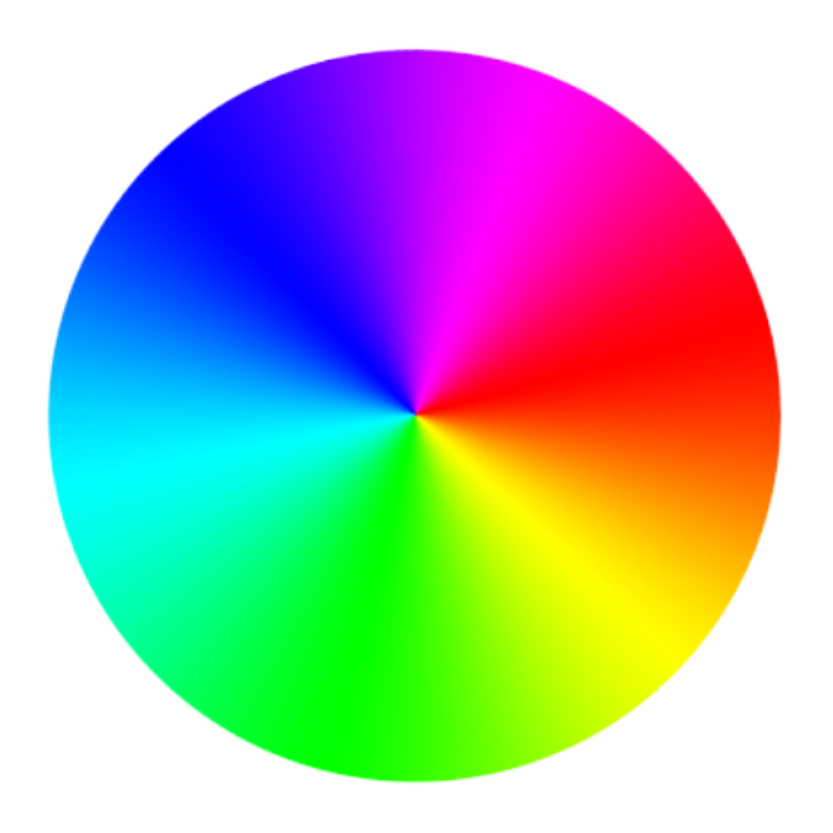}\\
Frame~50 & \includegraphics[width=0.3\textwidth]{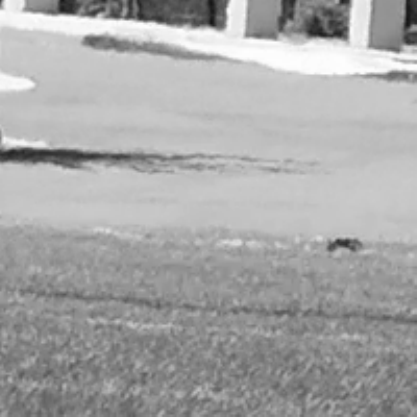} & \includegraphics[width=0.3\textwidth]{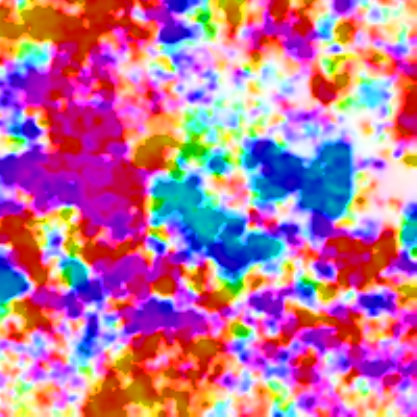} &
\end{tabular}
\caption{Several frames from the OTIS Car1 sequence (frame index in first column and actual frame in second column). The third column shows the velocity vector field corresponding to each frame.}
\label{fig:seq1}
\end{figure}

\begin{figure}[!h]
\centering
\begin{tabular}{m{0.15\textwidth}m{0.3\textwidth}m{0.3\textwidth}m{0.1\textwidth}}
Frame 150 & \includegraphics[width=0.3\textwidth]{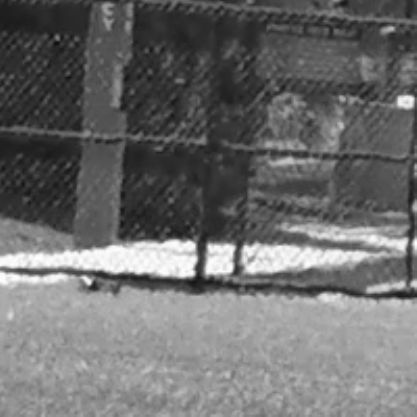} & \includegraphics[width=0.3\textwidth]{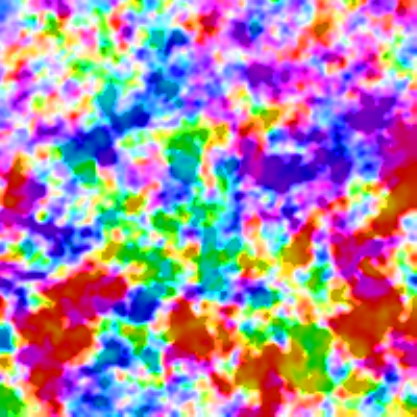} & \\
Frame 200 & \includegraphics[width=0.3\textwidth]{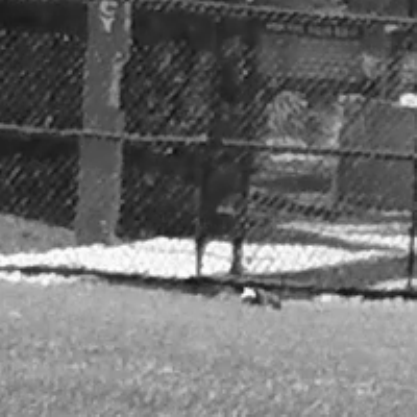} & \includegraphics[width=0.3\textwidth]{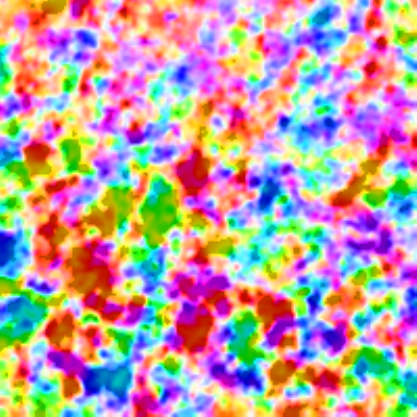} & \includegraphics[scale=0.07]{colorwheel} \\
Frame 250 & \includegraphics[width=0.3\textwidth]{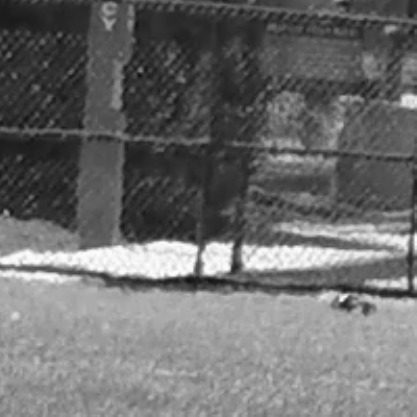} & \includegraphics[width=0.3\textwidth]{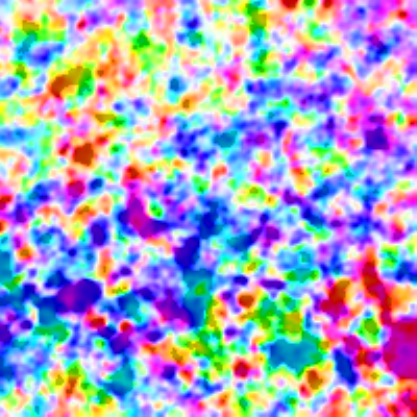} & 
\end{tabular}
\caption{Several frames from the OTIS Car2 sequence (frame index in first column and actual frame in second column). The third column shows the velocity vector field corresponding to each frame.}
\label{fig:seq2}
\end{figure}

\begin{figure}[!h]
\centering
\begin{tabular}{m{0.15\textwidth}m{0.3\textwidth}m{0.3\textwidth}m{0.1\textwidth}}
Frame 20 & \includegraphics[width=0.3\textwidth]{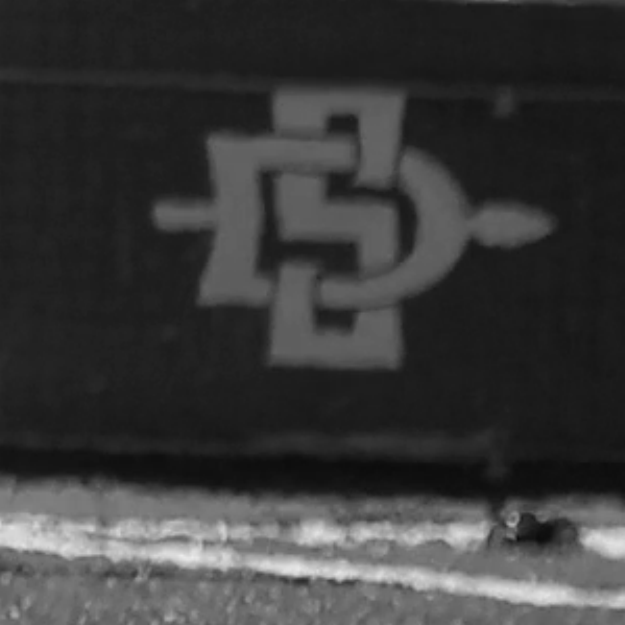} & \includegraphics[width=0.3\textwidth]{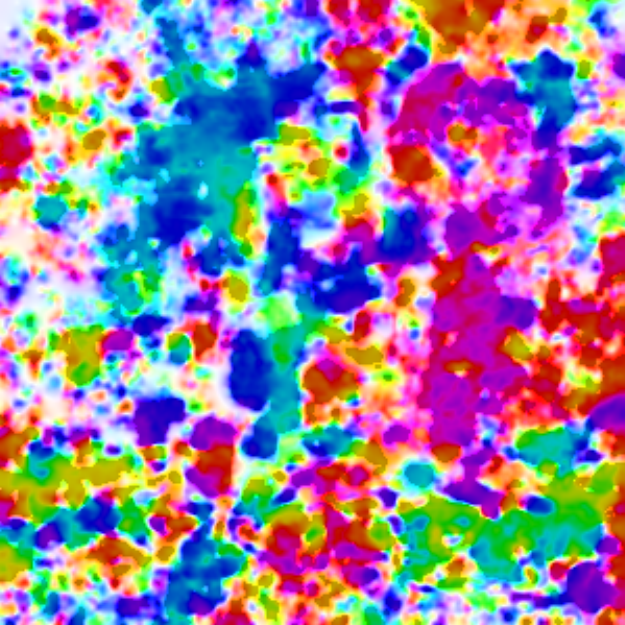} & \\
Frame 40 & \includegraphics[width=0.3\textwidth]{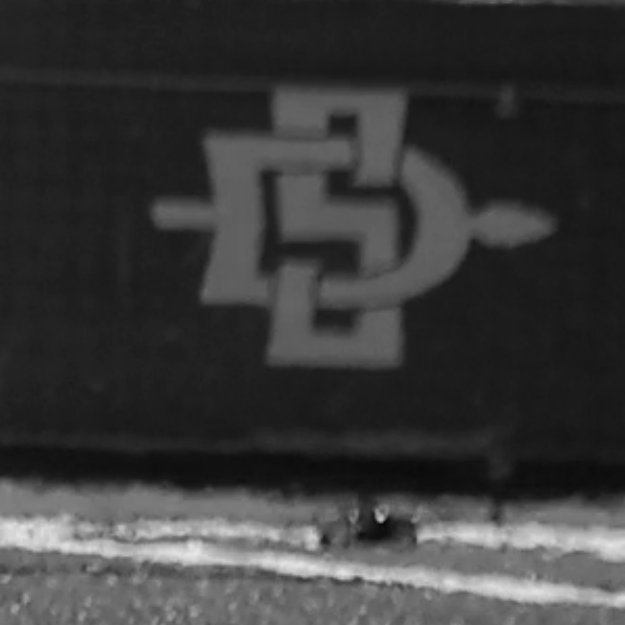} & \includegraphics[width=0.3\textwidth]{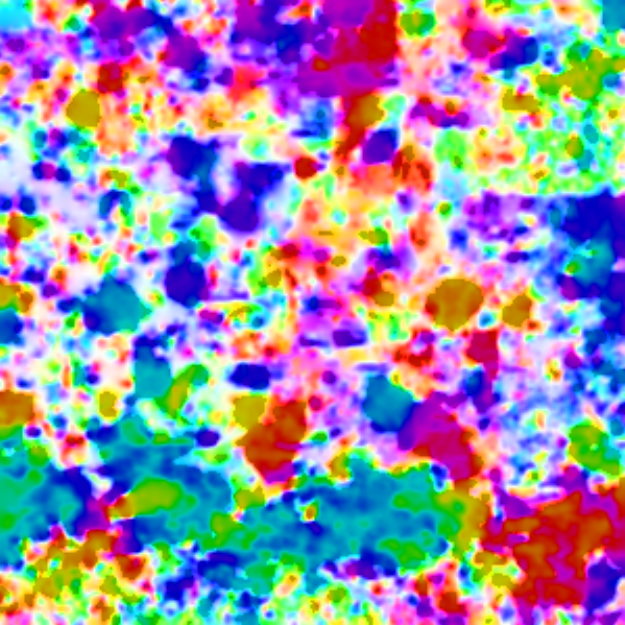} & \includegraphics[scale=0.07]{colorwheel} \\
Frame 60 & \includegraphics[width=0.3\textwidth]{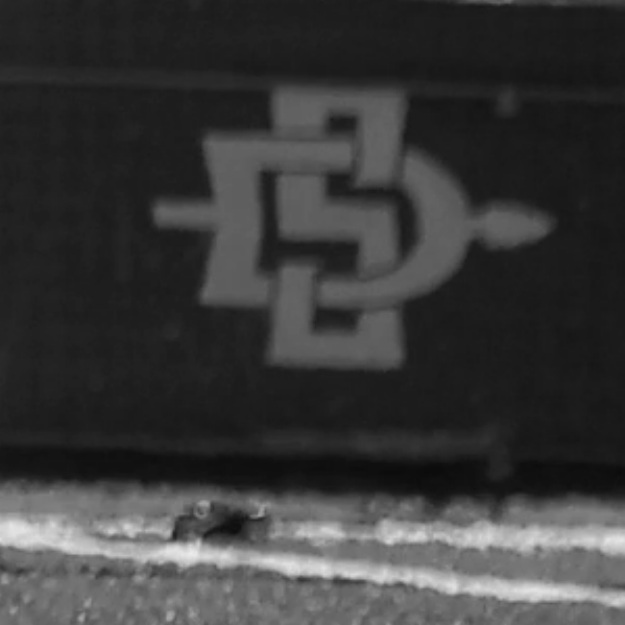} & \includegraphics[width=0.3\textwidth]{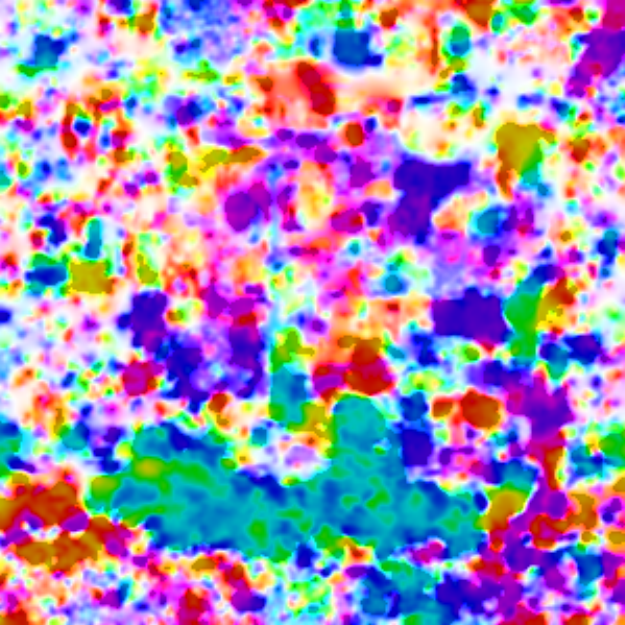} & 
\end{tabular}
\caption{Several frames from the OTIS Car4 sequence (frame index in first column and actual frame in second column). The third column shows the velocity vector field corresponding to each frame.}
\label{fig:seq3}
\end{figure}

\begin{figure}[!t]
\centering
\subfloat[Car1]{\includegraphics[width=0.45\textwidth]{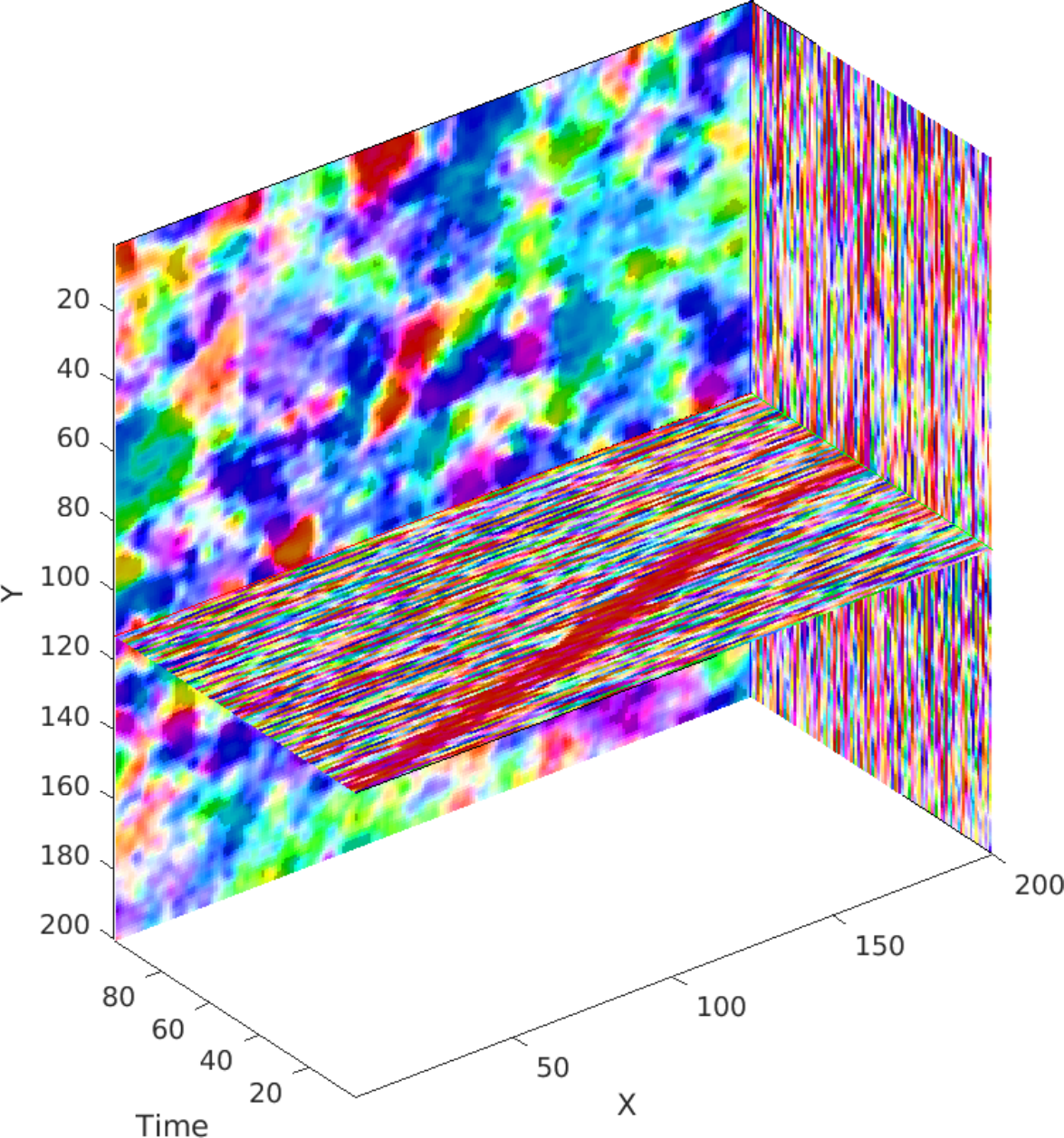}}\hspace{1.1mm}
\subfloat[Car2]{\includegraphics[width=0.49\textwidth]{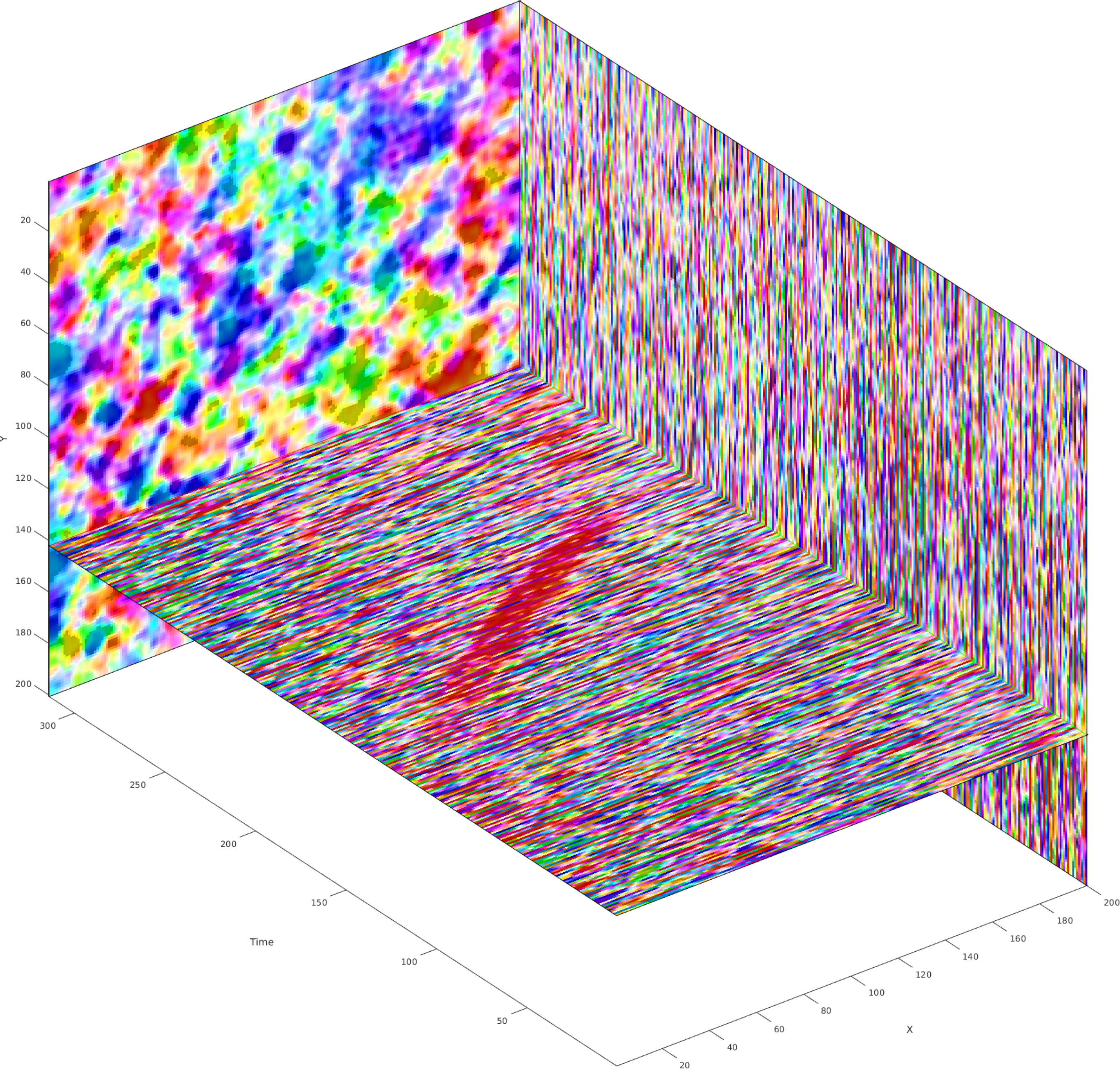}}
\caption{3D spatio-temporal patterns visualization in the velocity vector fields of the Car1 and Car2 sequences.}
\label{fig:spatiotime}
\end{figure}

Let us first illustrate some specific sequences impacted by the presence of atmospheric turbulence. Figures~\ref{fig:seq1}, \ref{fig:seq2} and \ref{fig:seq3} present several 
frames from the sequences Car1, Car2 and Car4 (see movies car1.avi\footnote{\url{http://jegilles.sdsu.edu/movies/VFD/car1.avi}}, 
car2.avi\footnote{\url{http://jegilles.sdsu.edu/movies/VFD/car2.avi}} and car4.avi\footnote{\url{http://jegilles.sdsu.edu/movies/VFD/car4.avi}}, note that we didn't use the Car3 
sequence from OTIS since it is very similar to the Car4 sequence) from the Open 
Turbulent Image Set (OTIS) \cite{OTIS}. Here we used the Diffeomorphic Demons algorithm to estimate the deformation flow. Of course other flow estimation techniques could be used and we investigate the influence of choosing other algorithms in Section~\ref{sec:exp}. Clearly, the turbulence creates dynamic local deformations which are confirmed 
by the velocity vector fields given on the second column (see movies car1-mov.avi\footnote{\url{http://jegilles.sdsu.edu/movies/VFD/car1-mov.avi}}, 
car2-mov.avi\footnote{\url{http://jegilles.sdsu.edu/movies/VFD/car2-mov.avi}} and car4-mov.avi\footnote{\url{http://jegilles.sdsu.edu/movies/VFD/car4-mov.avi}}) of 
each figure (the color wheel on the right of the figures gives the correspondence between the color and the vector direction). Each pixel in a given frame is clearly subject to some movement. 
Moreover, these vector fields primarily consist of random vectors (both in direction and magnitude). Therefore, it is almost impossible to distinguish the movement information 
which corresponds to moving object vs. moving atmosphere; therefore a new approach is needed. Our vision of the problem is biased by the fact that we watch the sequence frame by 
frame (i.e. a temporal succession of 2D images). In this paper, we consider the whole sequence as a single spatio-temporal space. In particular, if we visualize the colored vector 
field as a cube of data (i.e. each pixel in the 3D space is a 2D vector), we can see the emergence of geometric patterns. Indeed, if the turbulent movement is random in both space and 
time, the movement of a moving object is expected to be ``consistent'', i.e a moving object follows some non-erratic trajectory. This concept is illustrated in 
Figure~\ref{fig:spatiotime} where slices corresponding to $x-time$ planes ($y$ is fixed) are shown. This visualization confirms that a moving object creates a 
geometric pattern in the 
space-time volume while the atmospheric turbulence velocity field behaves in a random fashion. Although the classic Kolmogorov theory of turbulence shows that some autosimilarity 
distribution exists within the turbulence, such property is difficult to use to model the movement field induced by the turbulence. Instead, we will take a purely visual point of 
view: when we observe a sequence of frames, we can notice that fixed strutures in the images oscillate about their average position. This observation suggests that we can 
consider that components of the turbulence vector field can be modeled by oscillating functions. Therefore, we suggest that detecting moving objects through atmospheric 
turbulence can be achieved by separating non-oscillating vs. oscillating components of the vector fields. This idea is explored in the next section.

\section{Decomposition methods}\label{sec:dec}
The idea of separating oscillating vs. non-oscillating components has been widely studied in image processing. The specific purpose of these investigations was to separate the cartoon component from the texture component in a given image. The cartoon and texture parts correspond to non-oscillating and oscillating functions, respectively. In this section, we extend such algorithm to spatio-temporal vector fields. We first recall the formalism used in the image processing case and then introduce the vector field case.
\subsection{Image decomposition}
Given an image $f$, the purpose of the cartoon + texture decomposition is to find two images $u$ (the cartoon, i.e. piecewise constant function) and $v$ (the textures, i.e. oscillating function) such that $f=u+v$. This concept was proposed by Meyer \cite{meyer} in his investigation of the Rudin-Osher-Fatemi (ROF) model \cite{rof}. Meyer proposed to solve the following variational model to find such a decomposition.
$$(\hat{u},\hat{v})=\underset{u\in BV,v\in G}{\arg\min} \|u\|_{BV}+\lambda \|v\|_G,$$
where $\lambda$ is a parameter, $BV$ is the space of functions of bounded variation and $G$ is the dual space of the closure of $BV$ in the Schwartz class; this space $G$ corresponds to oscillating functions. Unfortunately, the norm on the $G$ space is difficult to handle numerically. Several authors \cite{aujol,aujoluvw,aujol2,vese2,starckimage,vese1} have proposed different modifications of this model, leading to tractable algorithms. In particular, Aujol et al. \cite{aujol} proposed to solve the following model.
\begin{equation}
(\hat{u},\hat{v})=\underset{u\in BV,v\in G_\mu}{\arg\min} \|u\|_{TV}+J_G^*\left(\frac{v}{\mu}\right)+(2\lambda)^{-1} \|f-(u+v)\|_{L^2},
\end{equation}
where $\lambda,\mu$ are parameters, $\|.\|_{TV}$ is the total variation (the norm associated with the homogeneous version of $BV$), $G_\mu=\{v\in G/ \|v\|_G\leq \mu\}$ and $J_G^*$ corresponds to the characteristic function on $G_\mu$ defined by
$$J_G^*\left(\frac{v}{\mu}\right)=\begin{cases} 0\qquad &\text{if}\quad v\in G_\mu\\ +\infty &\text{otherwise}\end{cases}.$$
Other function spaces were proposed in replacement of $BV$ and $G$. Sobolev spaces were used in \cite{vese2} and Besov spaces in \cite{aujoluvw}. The particular case of Besov spaces \cite{triebel2}, $\dot{B}_{p,q}^s$ (the homogeneous version of these spaces), is numerically interesting since Besov spaces can be characterized by the mean of wavelet coefficients \cite{chambolle2,daubechies,mallat}. Besov norms are defined by \eqref{eq:hdefbesov}.
\begin{equation}\label{eq:hdefbesov}
\forall f\in \dot{B}_{p,q}^s \qquad \|f\|_{\dot{B}_{p,q}^s}=\left(\sum_{j=-\infty}^{+\infty}2^{j\left(1-\frac{1}{p}+s\right)q}\left[\sum_{n\in\mathbb{Z}} 2^{j\frac{p}{2}}|\langle f,\psi_{jn}\rangle|^p\right]^{q/p}\right)^{1/q},
\end{equation}
where $\{\psi_{jn}\}$ is the set of wavelet functions ($j$ is the shifting factor and $n$ the scale). Using the fact that $\dot{B}_{1,1}^1\subset BV$ and $G\subset \dot{B}_{-1,\infty}^\infty$, we can formulate a Besov based variational decomposition model by
\begin{equation}
(\hat{u},\hat{v})=\underset{u\in \dot{B}_{1,1}^1,v\in \dot{B}_{-1,\infty}^\infty}{\arg\min} \|u\|_{\dot{B}_{1,1}^1}+J_B^*\left(\frac{v}{\mu}\right)+(2\lambda)^{-1} \|f-(u+v)\|_{L^2},
\end{equation}
where
\begin{equation}
J_B^*\left(\frac{v}{\mu}\right)=\begin{cases} 0\qquad &\text{if}\quad \|v\|_{\dot{B}_{-1,\infty}^\infty}\leq\mu\\ +\infty &\text{otherwise}\end{cases}.
\end{equation}
Such a model can easily be solved by iterating soft thresholding of the wavelet coefficients. Denoting $\mathcal{W}(f)$ as the wavelet transform of $f$ (we assume a real transform) and $Shrink(x,\lambda)=sign(x)\max(0,x-\lambda)$, the numerical algorithm can be resumed by Algorithm~\ref{alg:besov}.
\begin{algorithm}
\caption{Besov based decomposition numerical algorithm.}
\label{alg:besov}
\begin{algorithmic}
\STATE{Inputs: image to decompose $f$, parameters $\lambda,\mu$, maximum number of iterations $N_{max}$}
\STATE{Initialization: $n=0,u^0=v^0=0$}
\REPEAT
\STATE{$v^{n+1}=f-u^n-\mathcal{W}^{-1}\left(Shrink(\mathcal{W}(f-u^n),2\mu)\right)$}
\STATE{$u^{n+1}=\mathcal{W}^{-1}\left(Shrink(\mathcal{W}(f-v^{n+1}),2\lambda)\right)$}
\UNTIL{$\max\left(\|u^{n+1}-u^n\|_{L^2},\|v^{n+1}-v^n\|_{L^2}\right)<10^{-6}$ \OR $N_{max}$ iterations are reached}
\RETURN $u^{n+1},v^{n+1}$
\end{algorithmic}
\end{algorithm}

\subsection{Spatio-temporal vector field decomposition}
We now extend the previous decomposition model to spatio-temporal vector fields. We denote $f(i,j,n)$ the input sequence of $N$ frames where $i,j$ are the 2D spatial variables and $n$ is the frame number. The decomposition model can be easily extended to 3D by simply using a 3D wavelet transform instead of the 2D transform, while the algorithm itself remains unchanged. In this paper, we want to decompose vector fields corresponding to the estimated velocity from frame to frame; hence the previous algorithm must be adapted for vector fields. If we denote the spatio-temporal vector field $\mathbf{v}(i,j,n)=(v_1(i,j,n),v_2(i,j,n))$, the first idea that comes to mind is to process the vector field components $v_1$ and $v_2$ separately and then recompose the final vector field. Unfortunately, this approach does not perform well in some cases, like when one of the components is oscillating while the other is constant. We need to process the vector field as a single object. We can simply achieve this by rewriting the vector field as a complex vector field, i.e 
$$\tilde{\mathbf{v}}(i,j,n)=v_1(i,j,n)+\imath v_2(i,j,n).$$ 
Now, we can use the same 3D decomposition algorithm if we change the standard wavelet transform by a complex wavelet transform and then use the complex version of the thresholding operator defined by
$$\forall z=|z|e^{\imath\theta}\in\mathbb{C}\quad,\quad CShrink(z,\lambda)=\max(0,|z|-\lambda)e^{\imath\theta}.$$
We suggest a last modification to the algorithm to better take into account geometric patterns with specific orientations. As shown in Figure~\ref{fig:spatiotime}, the pattern associated to the moving object corresponds to a piecewise constant function with a direction not parallel to the space axis. Therefore, it is of interest to use a representation which takes into account such different orientations. The 3D classic wavelet transform is built on a tensor approach, i.e. is separable into three 1D transforms with respect to each axis. Hence the transform does not take orientations into consideration. An alternative to the classic wavelet transform is the curvelet transform \cite{candesfdct,contcurvelet1,contcurvelet2,candes99curvelets,donoho99digital,starck00curvelet,fdct3d} where the curvelet filters are built on angular wedges in the 3D Fourier space. Since curvelets are wavelet type functions, it is possible to formalize curvelet-based Besov spaces which we will denote $\dot{C}_{p,q}^s$. Norms on these spaces are defined by
\begin{equation}\label{eq:hdefcurv}
\forall f\in \dot{C}_{p,q}^s \qquad \|f\|_{\dot{C}_{p,q}^s}=\left(\sum_{j=-\infty}^{+\infty}2^{j\left(1-\frac{1}{p}+s\right)q}\left[\sum_{n\in\mathbb{Z}}\sum_{l=0}^{L_n} 2^{j\frac{p}{2}}|\langle f,\rho_{jln}\rangle|^p\right]^{q/p}\right)^{1/q},
\end{equation}
where $\{\rho_{jln}\}$ is the set of curvelet functions ($j$ is the shifting factor, $n$ the scale and $l$ the angular sector index where $L_n$ is the number of angular sectors at the scale $n$).
We can then replace the use of Besov spaces $\dot{B}_{p,q}^s$ by curvelet spaces $\dot{C}_{p,q}^q$ and the curvelet based decomposition model consists in solving model~\eqref{eq:curvfunc}.
\begin{equation}\label{eq:curvfunc}
(\hat{u},\hat{v})=\underset{u\in \dot{C}_{1,1}^1,v\in \dot{C}_{-1,\infty}^\infty}{\arg\min} \|u\|_{\dot{C}_{1,1}^1}+J_C^*\left(\frac{v}{\mu}\right)+(2\lambda)^{-1} \|f-(u+v)\|_{L^2},
\end{equation}
where
\begin{equation}
J_C^*\left(\frac{v}{\mu}\right)=\begin{cases} 0\qquad &\text{if}\quad \|v\|_{\dot{C}_{-1,\infty}^\infty}\leq\mu\\ +\infty &\text{otherwise}\end{cases}.
\end{equation}
If we denote $\mathcal{C}$ as the curvelet transform of $f$, the numerical algorithm that solves model~\eqref{eq:curvfunc} is given in Algorithm~\ref{alg:curvelet}.
\begin{algorithm}
\caption{Curvelet based decomposition numerical algorithm.}
\label{alg:curvelet}
\begin{algorithmic}
\STATE{Inputs: image to decompose $f$, parameters $\lambda,\mu$, maximum number of iterations $N_{max}$}
\STATE{Initialization: $n=0,u^0=v^0=0$}
\REPEAT
\STATE{$v^{n+1}=f-u^n-\mathcal{C}^{-1}\left(CShrink(\mathcal{C}(f-u^n),2\mu)\right)$}
\STATE{$u^{n+1}=\mathcal{C}^{-1}\left(CShrink(\mathcal{C}(f-v^{n+1}),2\lambda)\right)$}
\UNTIL{$\max\left(\|u^{n+1}-u^n\|_{L^2},\|v^{n+1}-v^n\|_{L^2}\right)<10^{-6}$ \OR $N_{max}$ iterations are reached}
\RETURN $u^{n+1},v^{n+1}$
\end{algorithmic}
\end{algorithm}

\section{Experimental results}\label{sec:exp}
\subsection{Decomposition results}
\begin{figure}[!h]
\centering
\begin{tabular}{m{2mm}m{0.3\textwidth}m{0.3\textwidth}m{0.1\textwidth}}
\rotatebox{90}{Frame~10} & \includegraphics[width=0.3\textwidth]{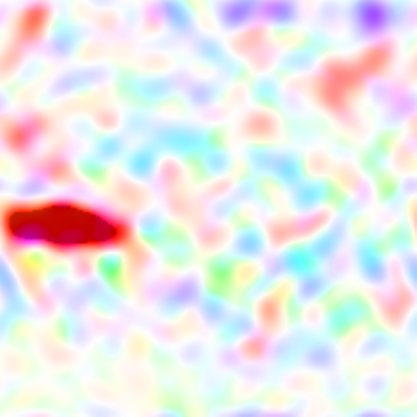} & \includegraphics[width=0.3\textwidth]{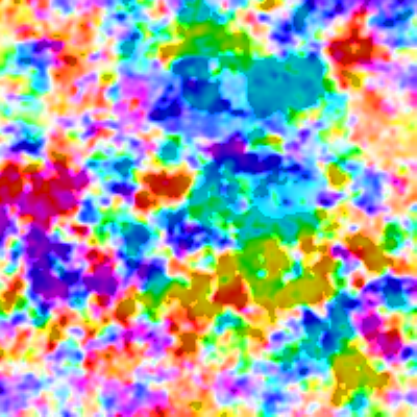} & \\
\rotatebox{90}{Frame~30} & \includegraphics[width=0.3\textwidth]{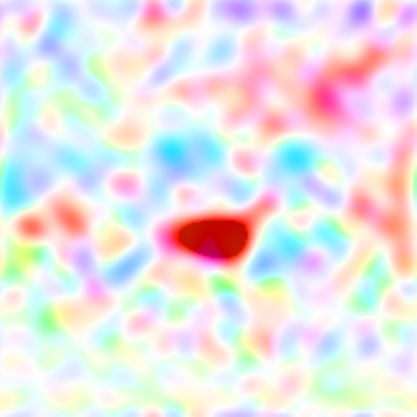} & \includegraphics[width=0.3\textwidth]{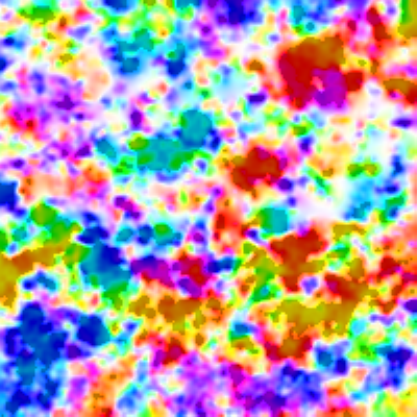} & 
\includegraphics[scale=0.07]{colorwheel}\\
\rotatebox{90}{Frame~50} & \includegraphics[width=0.3\textwidth]{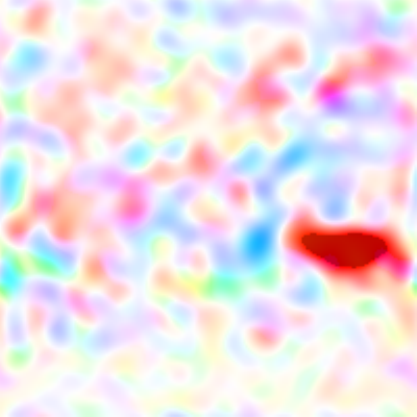} & \includegraphics[width=0.3\textwidth]{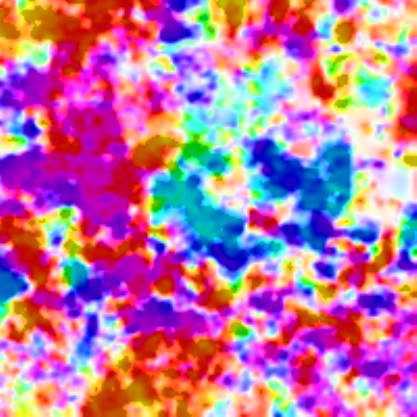} &
\end{tabular}
\caption{Decomposition results on the OTIS Car1 sequence. The first column gives the frame index, the second one corresponds to the geometric component and the third one to the oscillating component.}
\label{fig:UVseq1}
\end{figure}

\begin{figure}[!h]
\centering
\begin{tabular}{m{2mm}m{0.3\textwidth}m{0.3\textwidth}m{0.1\textwidth}}
\rotatebox{90}{Frame~150} & \includegraphics[width=0.3\textwidth]{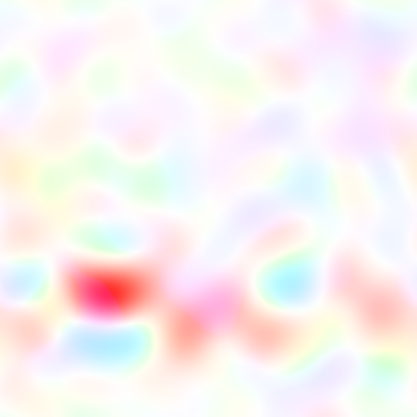} & \includegraphics[width=0.3\textwidth]{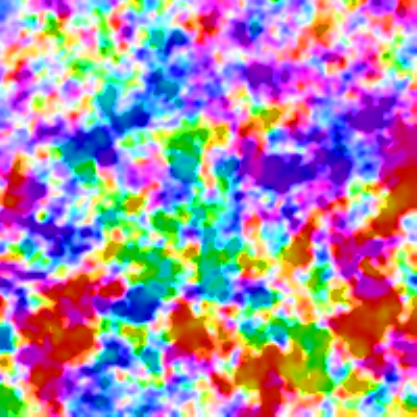} & \\
\rotatebox{90}{Frame~200} & \includegraphics[width=0.3\textwidth]{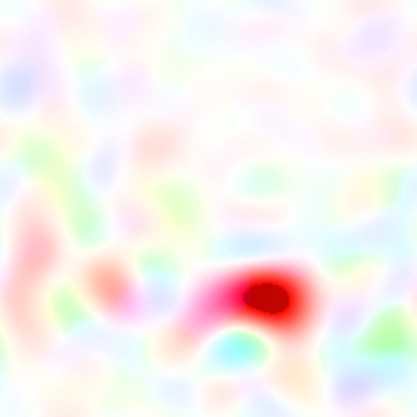} & \includegraphics[width=0.3\textwidth]{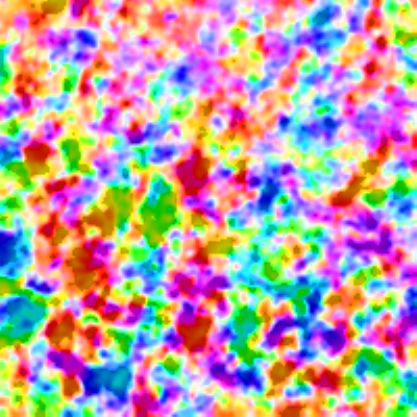} & 
\includegraphics[scale=0.07]{colorwheel}\\
\rotatebox{90}{Frame~250} & \includegraphics[width=0.3\textwidth]{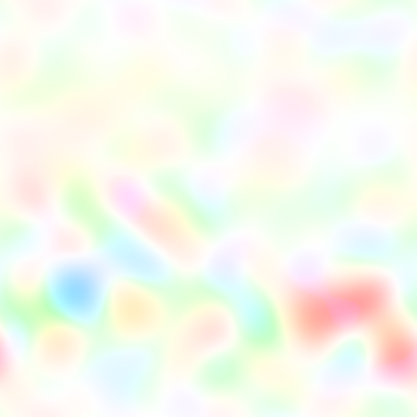} & \includegraphics[width=0.3\textwidth]{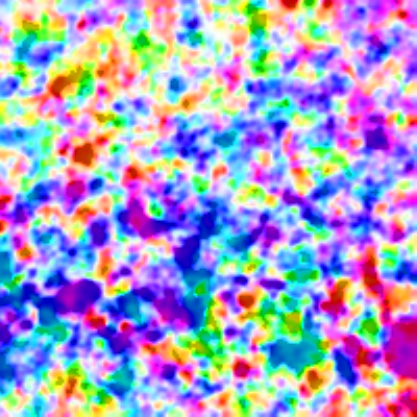} &
\end{tabular}
\caption{Decomposition results on the OTIS Car2 sequence. The first column gives the frame index, the second one corresponds to the geometric component and the third one to the oscillating component.}
\label{fig:UVseq2}
\end{figure}

\begin{figure}[!h]
\centering
\begin{tabular}{m{2mm}m{0.3\textwidth}m{0.3\textwidth}m{0.1\textwidth}}
\rotatebox{90}{Frame~20} & \includegraphics[width=0.3\textwidth]{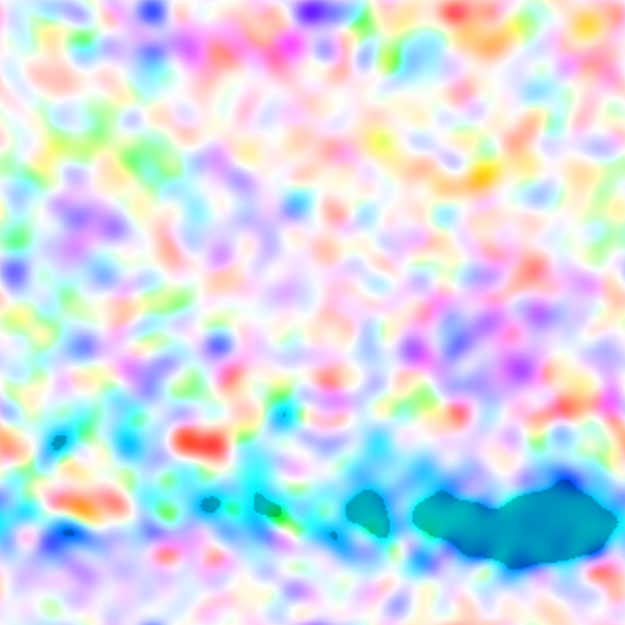} & \includegraphics[width=0.3\textwidth]{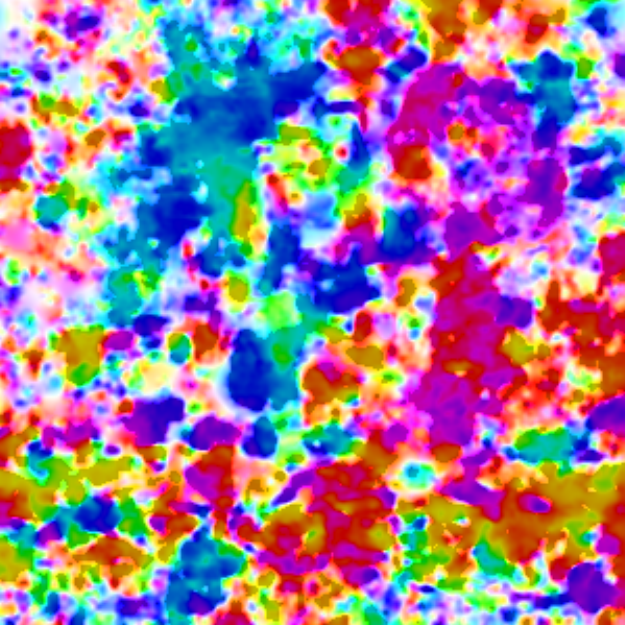} & \\
\rotatebox{90}{Frame~40} & \includegraphics[width=0.3\textwidth]{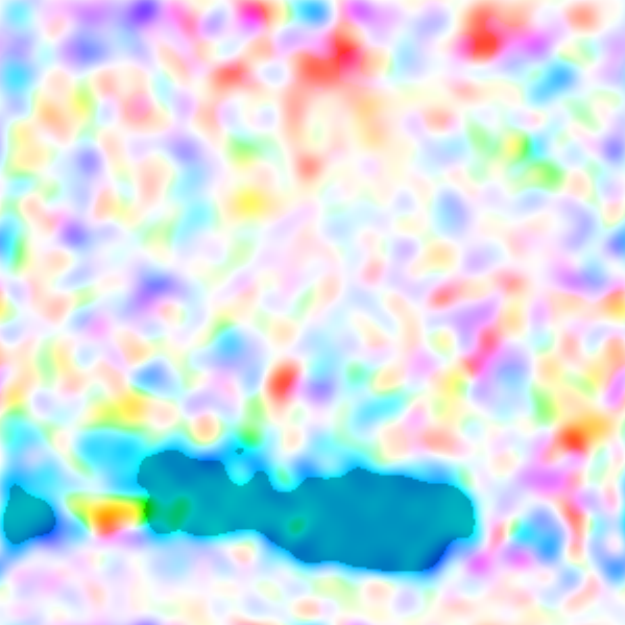} & \includegraphics[width=0.3\textwidth]{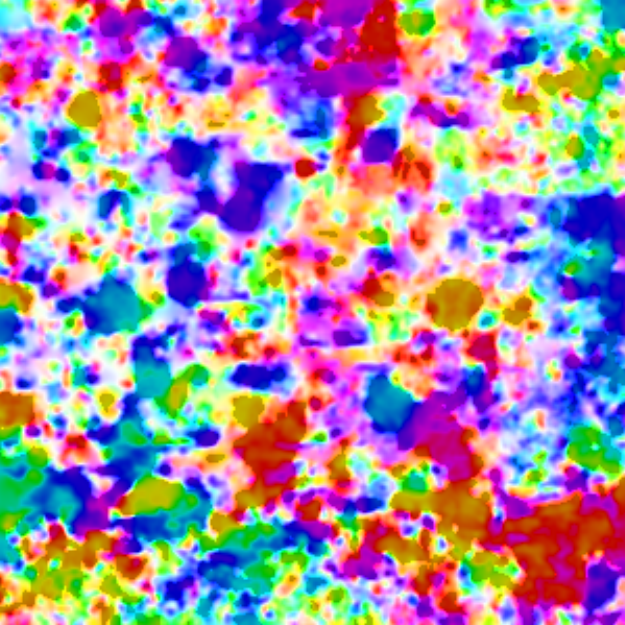} & 
\includegraphics[scale=0.07]{colorwheel}\\
\rotatebox{90}{Frame~60} & \includegraphics[width=0.3\textwidth]{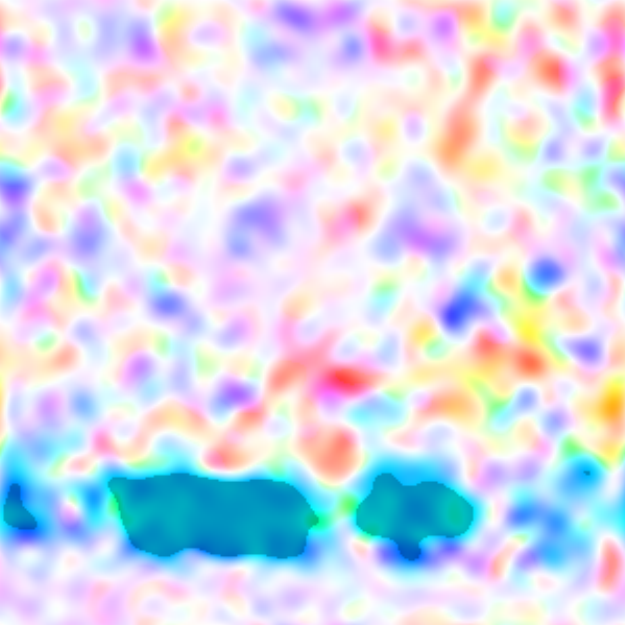} & \includegraphics[width=0.3\textwidth]{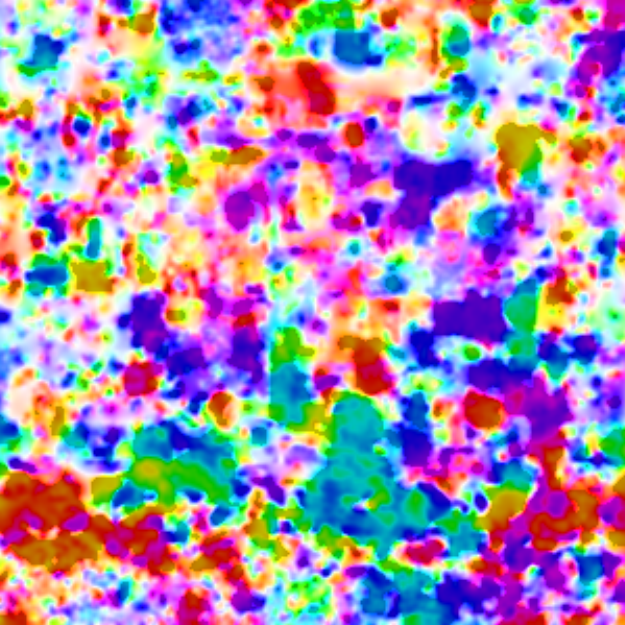} &
\end{tabular}
\caption{Decomposition results on the OTIS Car4 sequence. The first column gives the frame index, the second one corresponds to the geometric component and the third one to the oscillating component.}
\label{fig:UVseq4}
\end{figure}

\begin{figure}[!h]
\centering
\subfloat[Car1 - Geometric component]{\includegraphics[width=0.45\textwidth,height=6cm]{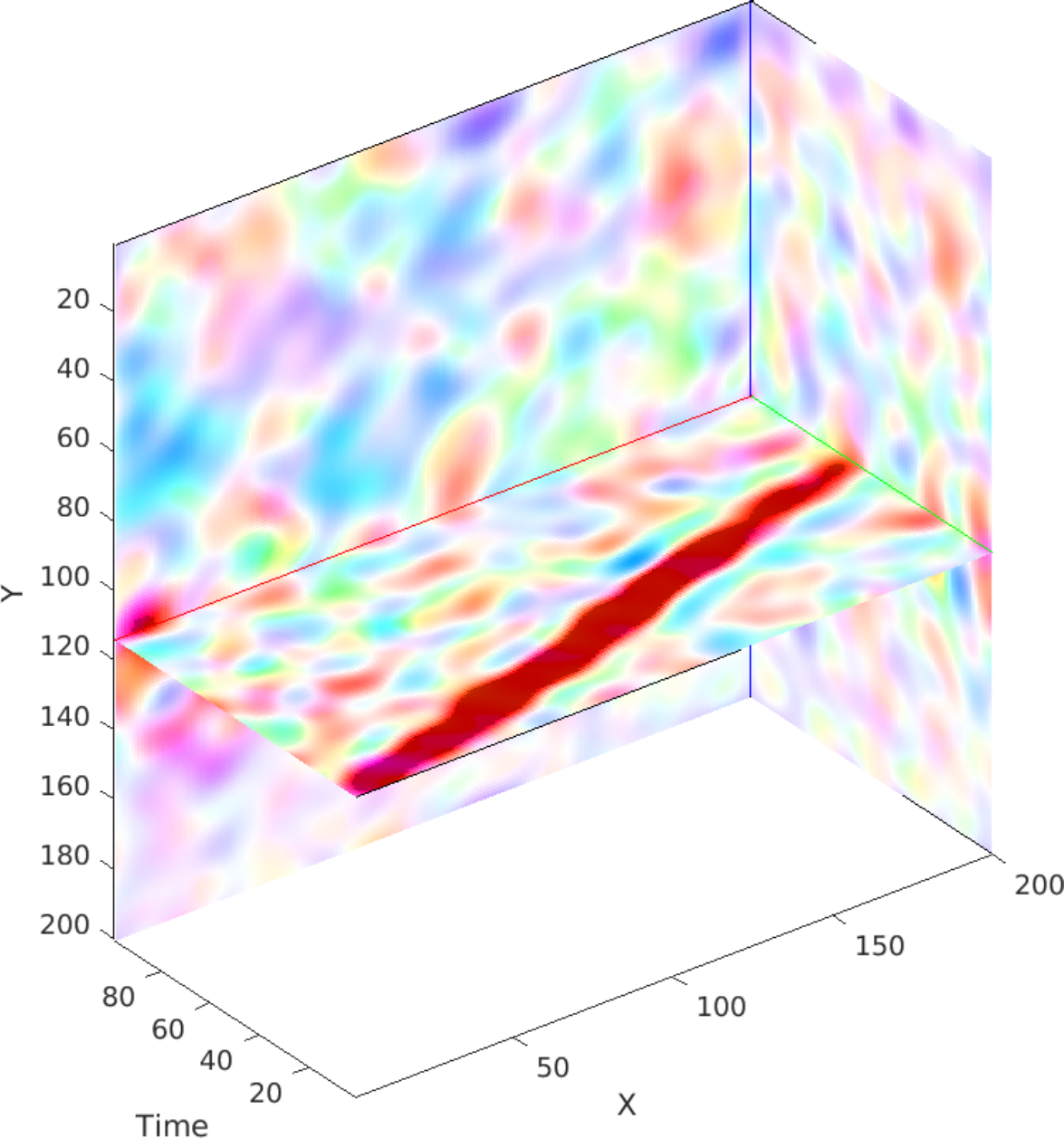}}\hspace{1mm}
\subfloat[Car1 - Oscillating component]{\includegraphics[width=0.45\textwidth,height=6cm]{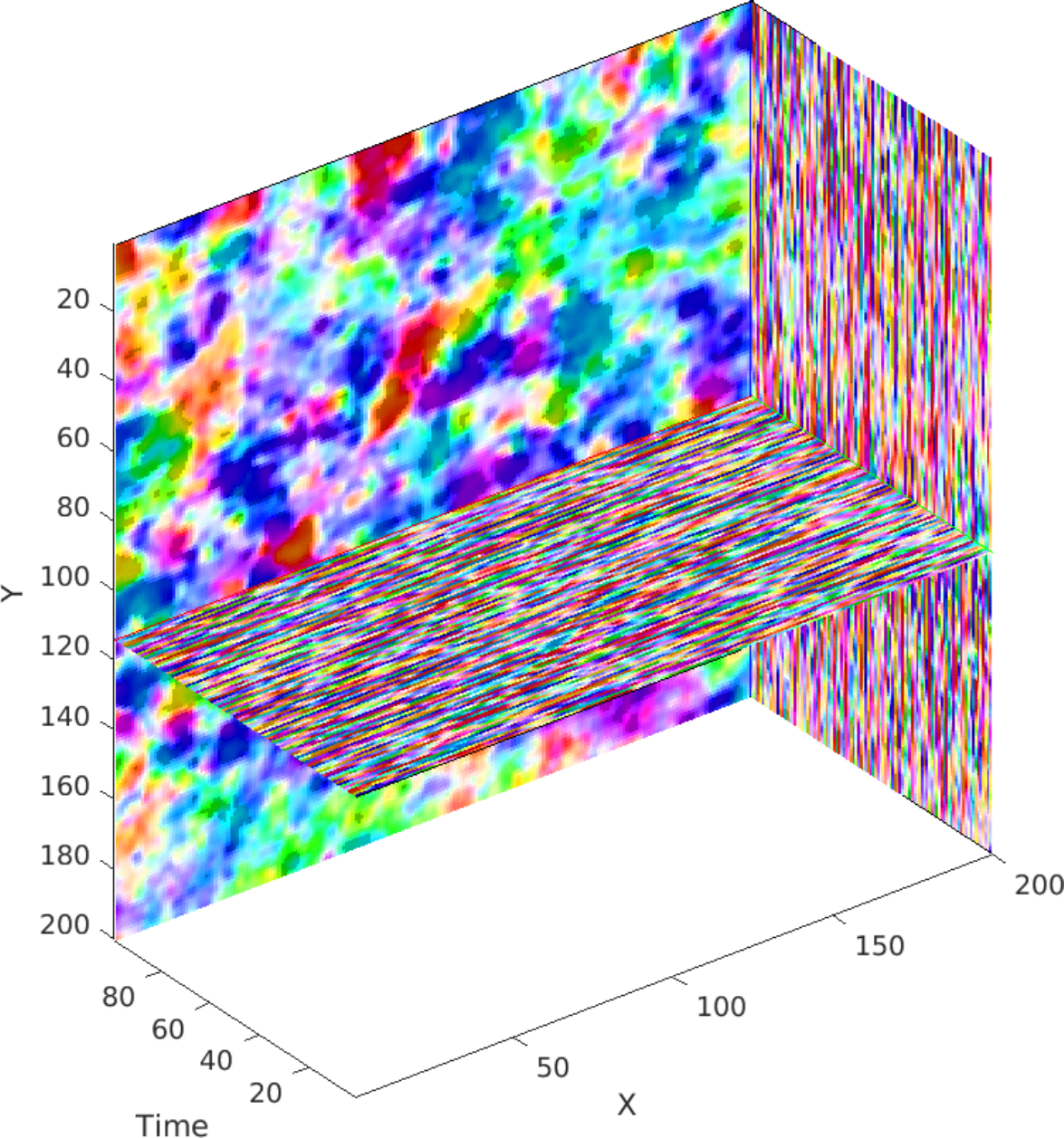}}\\
\subfloat[Car2 - Geometric component]{\includegraphics[width=0.45\textwidth,height=4cm]{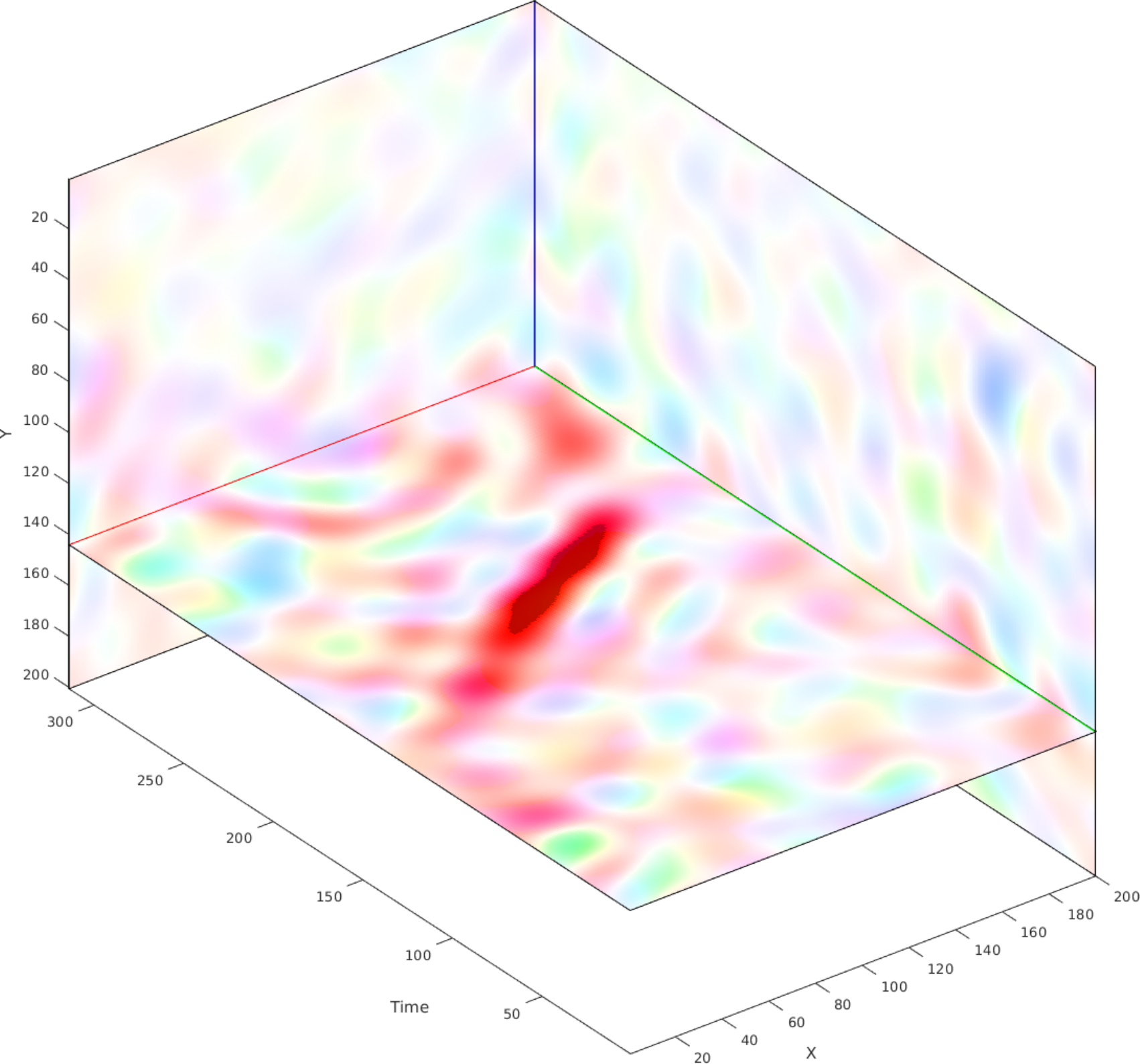}}\hspace{1mm}
\subfloat[Car2 - Oscillating component]{\includegraphics[width=0.45\textwidth,height=4cm]{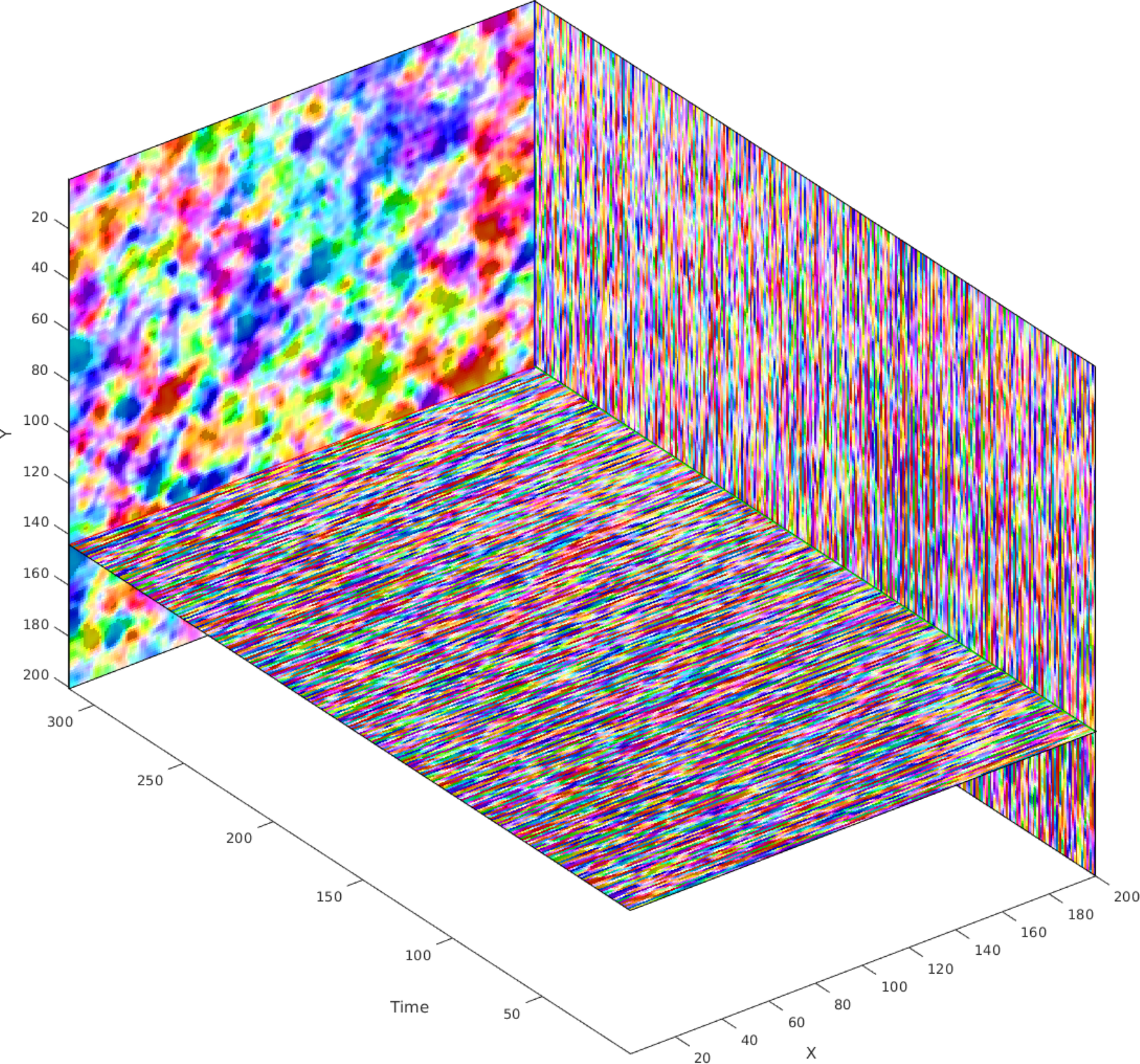}}
\caption{3D spatio-temporal patterns visualization of the geometric and oscillating components of the Car1 and Car2 sequences velocity vector fields.}
\label{fig:spatiotimeUV}
\end{figure}

In this section, we present the results obtained from Algorithm~\ref{alg:curvelet} on the sequences Car1, Car2 and Car4 from the OTIS dataset \cite{OTIS}. The algorithm was 
implemented in Matlab and used the curvelet transform provided in the Curvelab toolbox \cite{curvelab}. Regarding the choice of the parameters $\lambda$ and $\mu$, an 
automatic estimation of these parameters is still an open problem. We experimentally found that the choice $\lambda=\mu=1$ works well (and the final results were not sensitive to 
small variations of these parameters) for 
all sequences used in this work (we want emphasize that, a priori, this choice depends on the characteristics of the used images and may be adapted from one imaging system or 
acquisition conditions to another) and hence was used in all our experiments (we believe that this choice works because the ``geometric flows'' and ``oscillating flows'' are very 
well separated, i.e 
their characteristics are very different). The maximum number of iterations was fixed to $N_{max}=5$ (this choice was made on our experience in 2D cartoon+textures decomposition 
models as well as to keep the algorithm computationally tractable). Figures~\ref{fig:UVseq1}, \ref{fig:UVseq2} and \ref{fig:UVseq4} 
show the geometric and oscillating components of the velocity vector fields corresponding to each sequence (see the corresponding movies 
car1-dec.avi\footnote{\url{http://jegilles.sdsu.edu/movies/VFD/car1-dec.avi}}, car2-dec.avi\footnote{\url{http://jegilles.sdsu.edu/movies/VFD/car2-dec.avi}} and 
car4-dec.avi\footnote{\url{http://jegilles.sdsu.edu/movies/VFD/car4-dec.avi}} in the supplementary files), respectively. As expected, the geometric component highlighted the 
movement of the moving object while the oscillating component contained the turbulent movement. Even in the Car2 sequence (in which a small object moves close to the frame edge and 
is strongly affected by the turbulence), the algorithm performs well. For comparison with Figure~\ref{fig:spatiotime}, we provide the spatio-temporal plots for the extracted 
components in Figure~\ref{fig:spatiotimeUV}. Again, it is obvious to see that the expected geometric patterns corresponding to moving objects are enhanced in the geometric 
component, making them easier to detect.

\subsection{Impact on the detection}
\begin{figure}[!t]
\centering
\begin{tabular}{m{2mm}m{0.25\textwidth}m{0.25\textwidth}m{0.25\textwidth}}
\rotatebox{90}{Frame 10} & \includegraphics[width=0.25\textwidth]{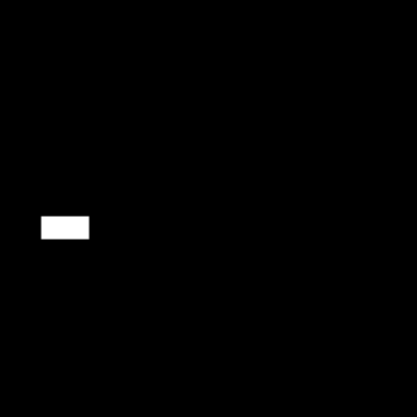} & 
\includegraphics[width=0.25\textwidth]{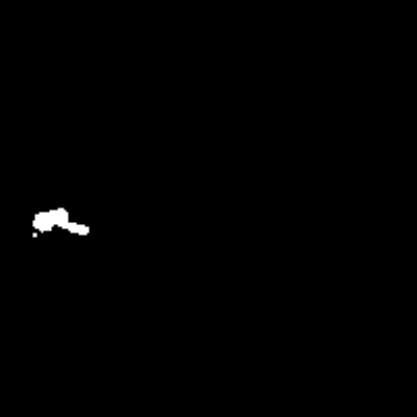} & \includegraphics[width=0.25\textwidth]{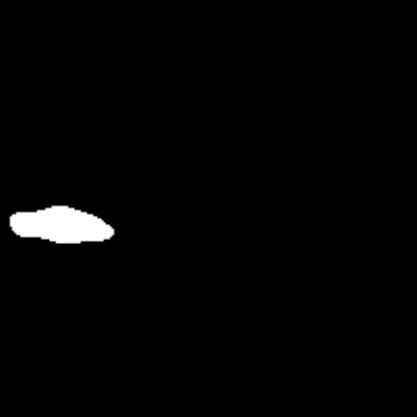} \\
\rotatebox{90}{Frame 30} & \includegraphics[width=0.25\textwidth]{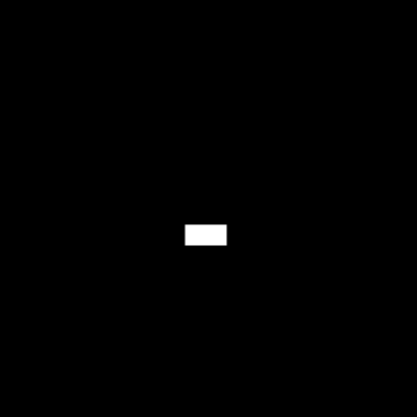} & \includegraphics[width=0.25\textwidth]{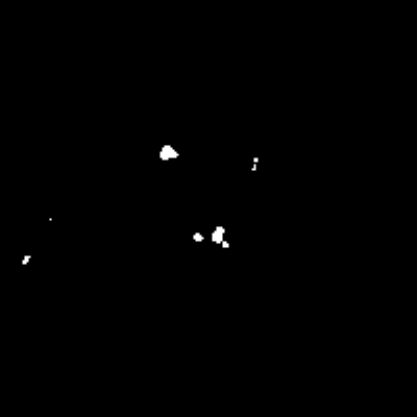} 
& \includegraphics[width=0.25\textwidth]{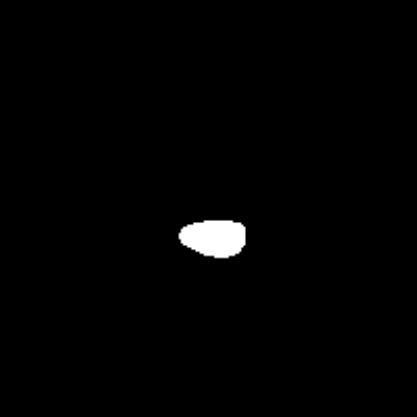} \\
\end{tabular}
\caption{Detection results on the Car1 sequence using the Demons flow. The left column is the groundtruth, the middle and right columns give the detected object from the 
original movement flow and geometric component of the flow, respectively.}
\label{fig:detectcar1}
\end{figure}

\begin{figure}[!t]
\centering
\begin{tabular}{m{2mm}m{0.25\textwidth}m{0.25\textwidth}m{0.25\textwidth}}
\rotatebox{90}{Frame 150} & \includegraphics[width=0.25\textwidth]{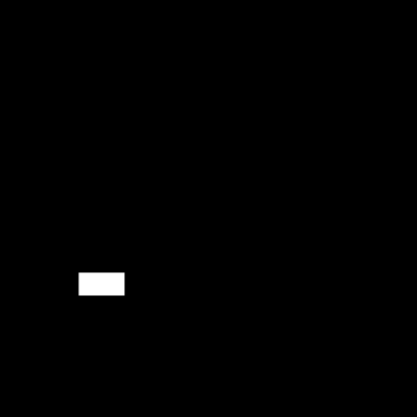} & 
\includegraphics[width=0.25\textwidth]{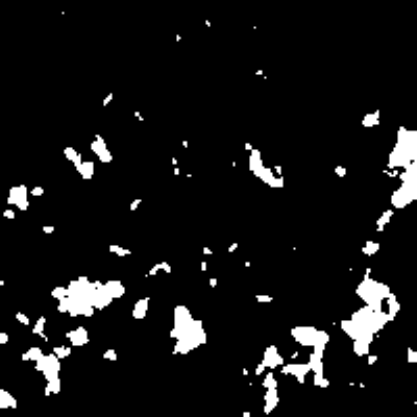} & \includegraphics[width=0.25\textwidth]{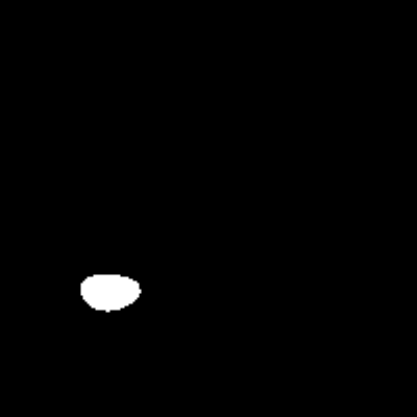} \\
\rotatebox{90}{Frame 200} & \includegraphics[width=0.25\textwidth]{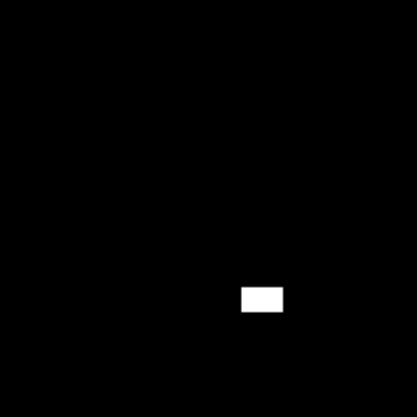} & 
\includegraphics[width=0.25\textwidth]{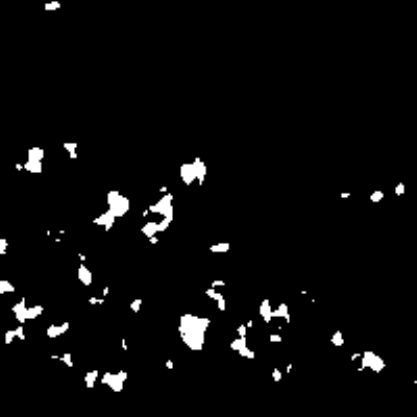} 
& \includegraphics[width=0.25\textwidth]{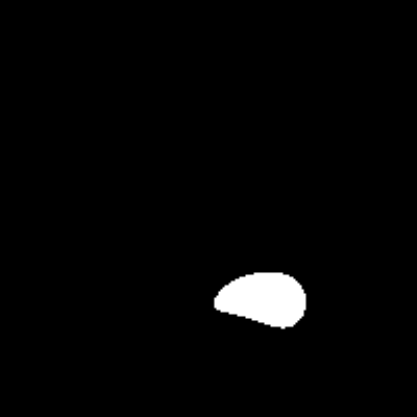}\\
\end{tabular}
\caption{Detection results on the Car2 sequence using the Demons flow. The left column is the groundtruth, the middle and right columns give the detected 
object from the original movement flow 
and geometric component of the flow, respectively.}
\label{fig:detectcar2}
\end{figure}

\begin{figure}[!t]
\centering
\begin{tabular}{m{2mm}m{0.25\textwidth}m{0.25\textwidth}m{0.25\textwidth}}
\rotatebox{90}{Frame 20} & \includegraphics[width=0.25\textwidth]{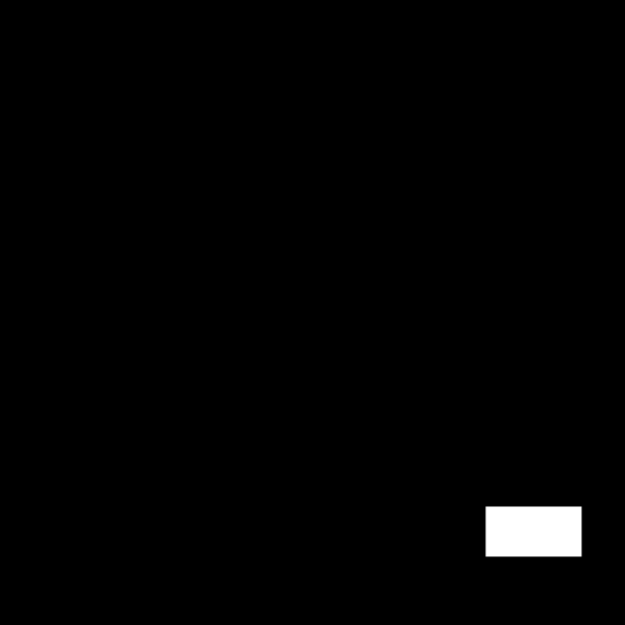} & 
\includegraphics[width=0.25\textwidth]{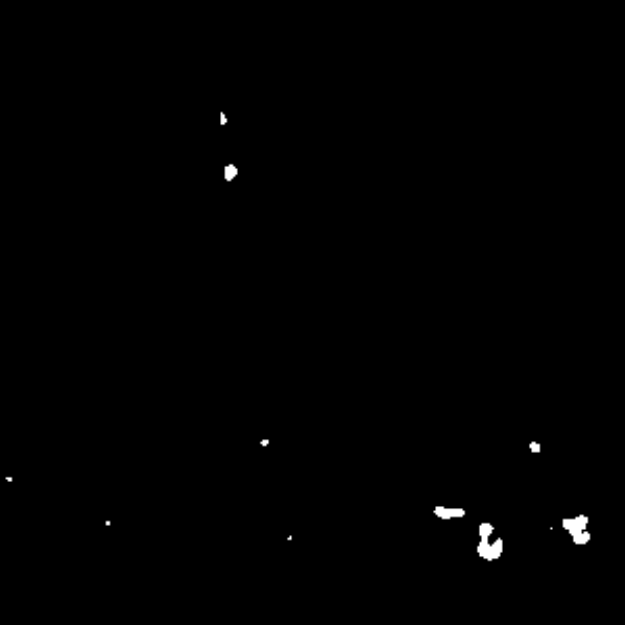} & \includegraphics[width=0.25\textwidth]{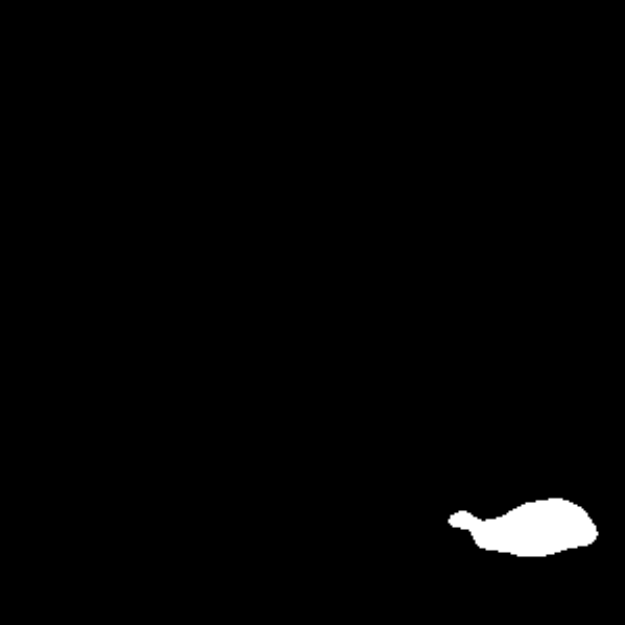} \\
\rotatebox{90}{Frame 40} & \includegraphics[width=0.25\textwidth]{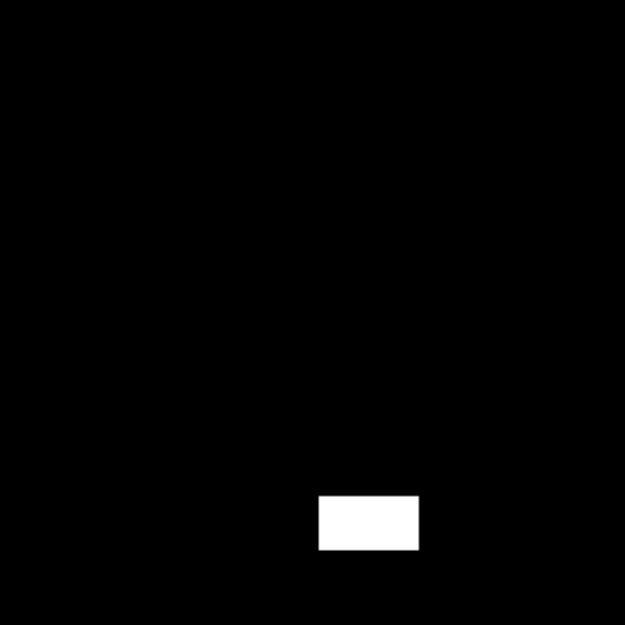} & \includegraphics[width=0.25\textwidth]{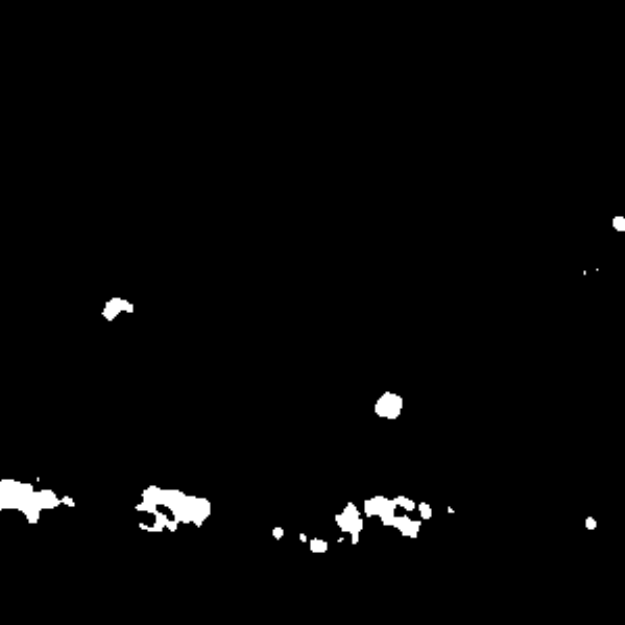} 
& \includegraphics[width=0.25\textwidth]{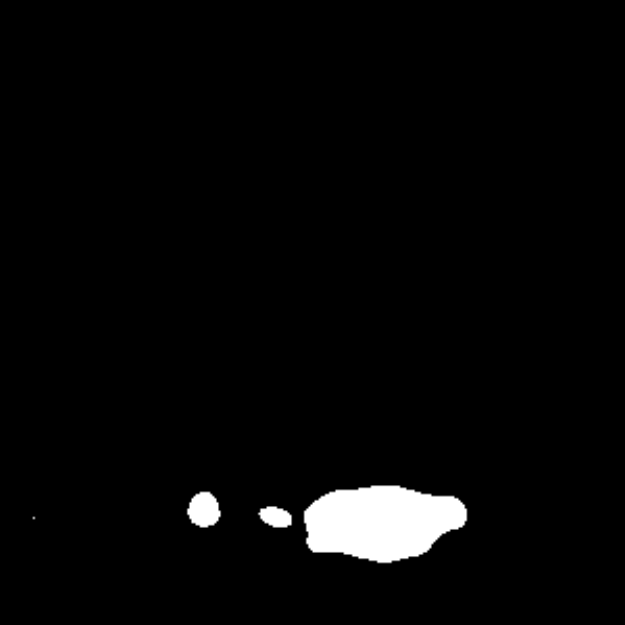}\\
\end{tabular}
\caption{Detection results on the Car4 sequence using the Demons flow. The left column is the groundtruth, the middle and right columns give the detected object from the 
original movement flow 
and geometric component of the flow, respectively.}
\label{fig:detectcar4}
\end{figure}

Even if the purpose of this paper is not to focus on the detection and tracking steps, in this section we compare detections obtained from both the original flow and from the 
geometric component of the flow. To do so we perform a simple detection technique which consists in computing the magnitude of the vector field and then threshold it. We 
manually set the thresholds in order to keep enough pixels from the moving object while removing a maximum of outliers.
Figures~\ref{fig:detectcar1}, \ref{fig:detectcar2} and \ref{fig:detectcar4} present the obtained results, as well as the groundtruths, on the Car1, Car2 and Car4 sequences respectively. The 
Car1 and Car2 sequences are the easiest ones and only a few outliers appear from the original flow. However, even if the size seems to be over estimated, only one object is 
detected from the geometric component of the flow for both sequences. The Car2 sequence is the most challenging one since the target velocity is close to the turbulence velocity 
and the target size is pretty small. If many outliers appear when using the original flow, the proposed technique is capable of getting rid of these outliers and only keep the expected moving object. 

\subsection{Influence of the used movement flow estimation technique}
\begin{figure}[!t]
\centering
\begin{tabular}{m{2mm}m{0.25\textwidth}m{0.25\textwidth}m{0.25\textwidth}}
\rotatebox{90}{Frame 10} & \includegraphics[width=0.25\textwidth]{car1GT10} & 
\includegraphics[width=0.25\textwidth]{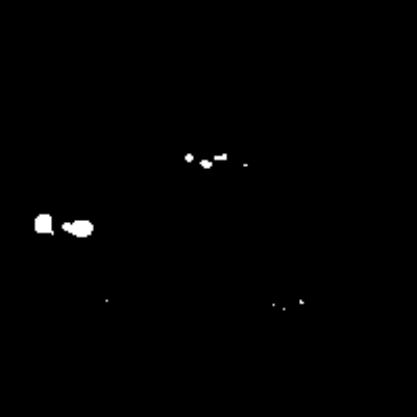} & \includegraphics[width=0.25\textwidth]{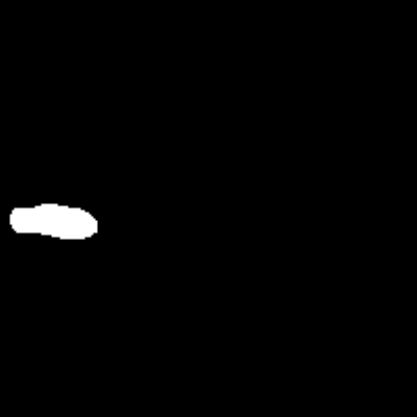} \\
\rotatebox{90}{Frame 30} & \includegraphics[width=0.25\textwidth]{car1GT30} & \includegraphics[width=0.25\textwidth]{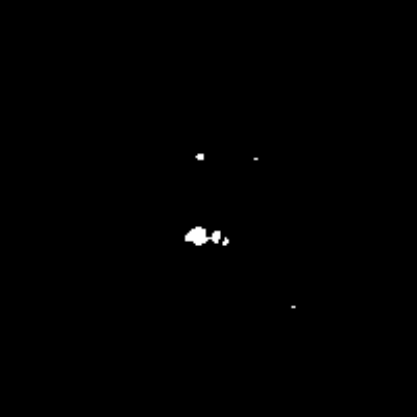} 
& \includegraphics[width=0.25\textwidth]{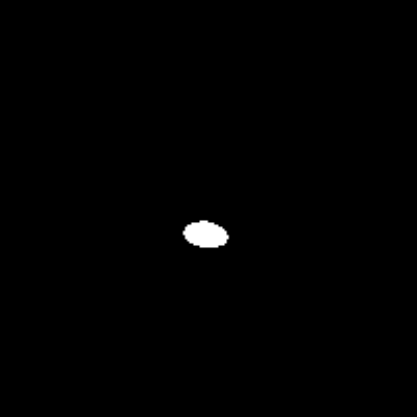}\\
\end{tabular}
\caption{Detection results on the Car1 sequence using the Horn-Schunck optical flow. The left column is the groundtruth, the middle and right columns give the detected 
object from the original movement flow 
and geometric component of the flow, respectively.}
\label{fig:hsdetectcar1}
\end{figure}

\begin{figure}[!t]
\centering
\begin{tabular}{m{2mm}m{0.25\textwidth}m{0.25\textwidth}m{0.25\textwidth}}
\rotatebox{90}{Frame 150} & \includegraphics[width=0.25\textwidth]{car2GT150} & 
\includegraphics[width=0.25\textwidth]{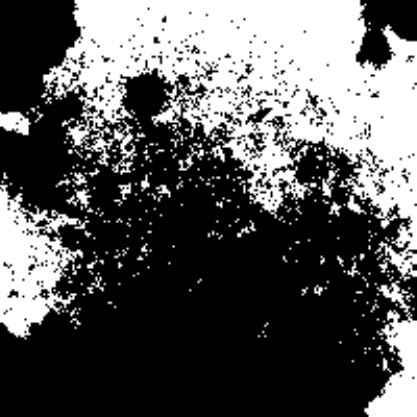} & \includegraphics[width=0.25\textwidth]{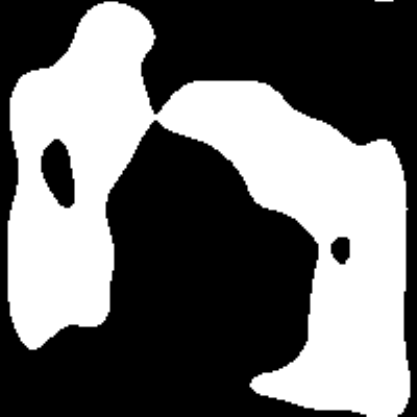} \\
\rotatebox{90}{Frame 200} & \includegraphics[width=0.25\textwidth]{car2GT200} & 
\includegraphics[width=0.25\textwidth]{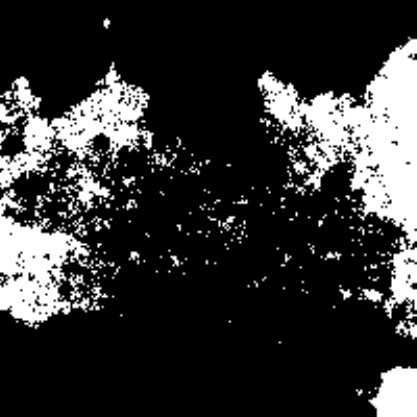} 
& \includegraphics[width=0.25\textwidth]{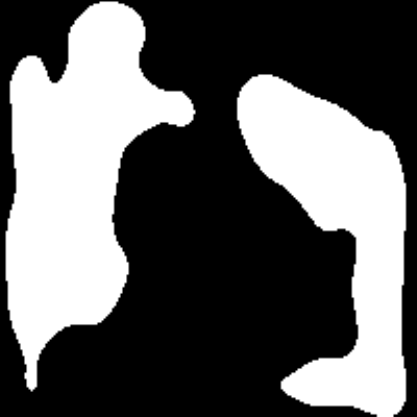}\\
\end{tabular}
\caption{Detection results on the Car2 sequence using the Horn-Schunck optical flow. The left column is the groundtruth, the middle and right columns give the detected 
object from the original movement flow 
and geometric component of the flow, respectively.}
\label{fig:hsdetectcar2}
\end{figure}

\begin{figure}[!t]
\centering
\begin{tabular}{m{2mm}m{0.25\textwidth}m{0.25\textwidth}m{0.25\textwidth}}
\rotatebox{90}{Frame 20} & \includegraphics[width=0.25\textwidth]{car4GT20} & 
\includegraphics[width=0.25\textwidth]{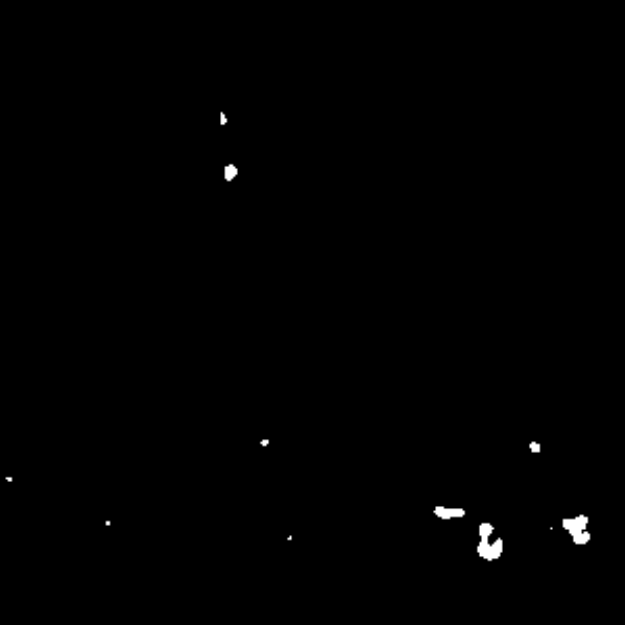} & \includegraphics[width=0.25\textwidth]{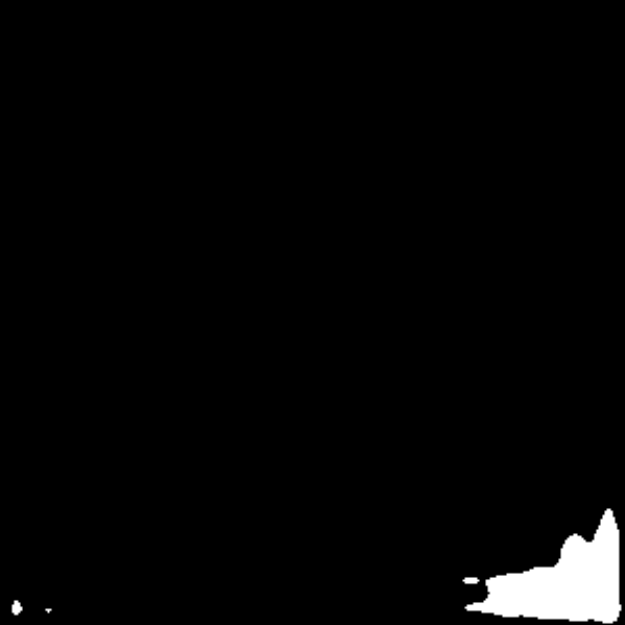} \\
\rotatebox{90}{Frame 40} & \includegraphics[width=0.25\textwidth]{car4GT40} & \includegraphics[width=0.25\textwidth]{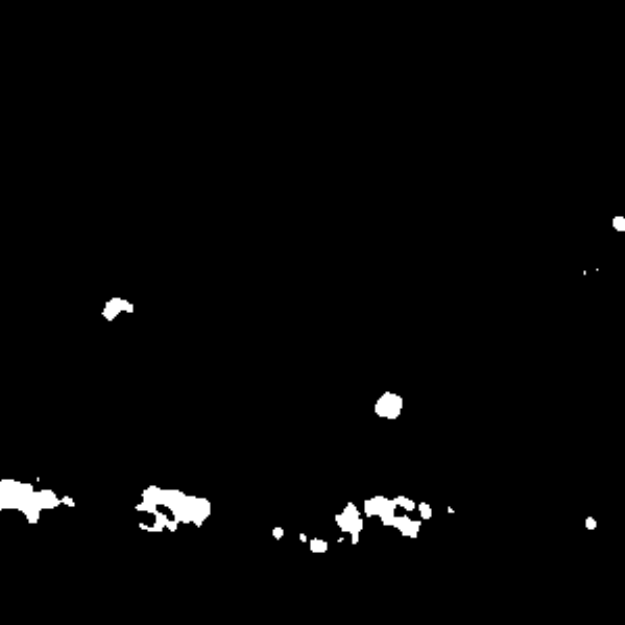} 
& \includegraphics[width=0.25\textwidth]{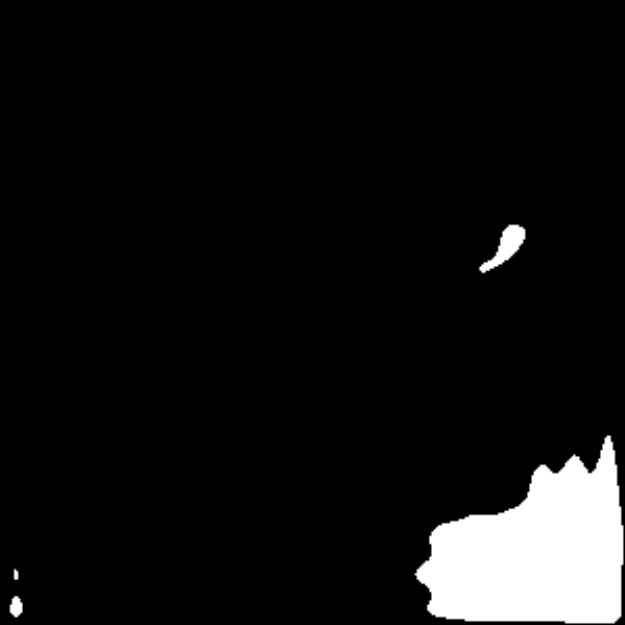}\\
\end{tabular}
\caption{Detection results on the Car4 sequence using the Horn-Schunck optical flow. The left column is the groundtruth, the middle and right columns give the detected 
object from the original movement flow 
and geometric component of the flow, respectively.}
\label{fig:hsdetectcar4}
\end{figure}

\begin{figure}[!t]
\centering
\begin{tabular}{m{2mm}m{0.25\textwidth}m{0.25\textwidth}m{0.25\textwidth}}
\rotatebox{90}{Frame 10} & \includegraphics[width=0.25\textwidth]{car1GT10} & 
\includegraphics[width=0.25\textwidth]{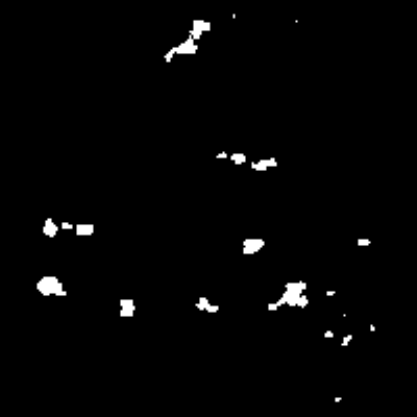} & \includegraphics[width=0.25\textwidth]{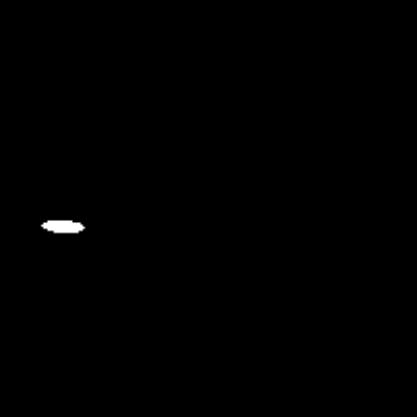} \\
\rotatebox{90}{Frame 30} & \includegraphics[width=0.25\textwidth]{car1GT30} & \includegraphics[width=0.25\textwidth]{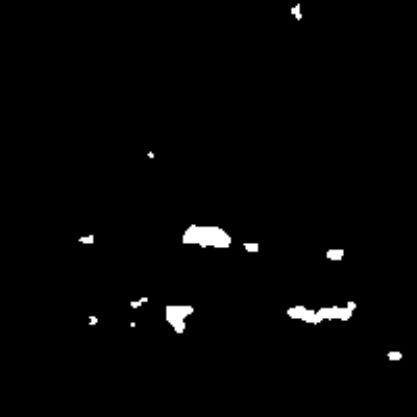} 
& \includegraphics[width=0.25\textwidth]{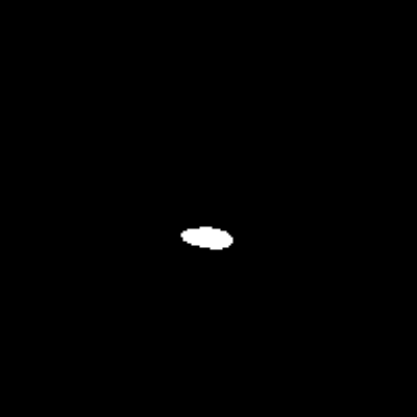}\\
\end{tabular}
\caption{Detection results on the Car1 sequence using the $TV-L^1$ optical flow. The left column is the groundtruth, the middle and right columns give the detected 
object from the original movement flow 
and geometric component of the flow, respectively.}
\label{fig:tvdetectcar1}
\end{figure}

\begin{figure}[!t]
\centering
\begin{tabular}{m{2mm}m{0.25\textwidth}m{0.25\textwidth}m{0.25\textwidth}}
\rotatebox{90}{Frame 150} & \includegraphics[width=0.25\textwidth]{car2GT150} & 
\includegraphics[width=0.25\textwidth]{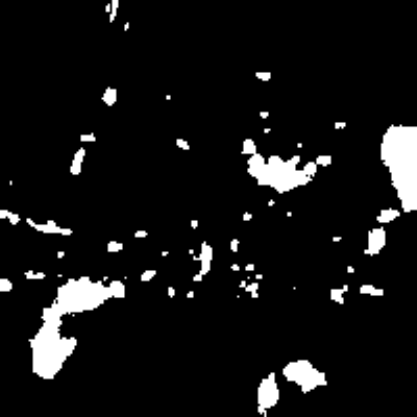} & \includegraphics[width=0.25\textwidth]{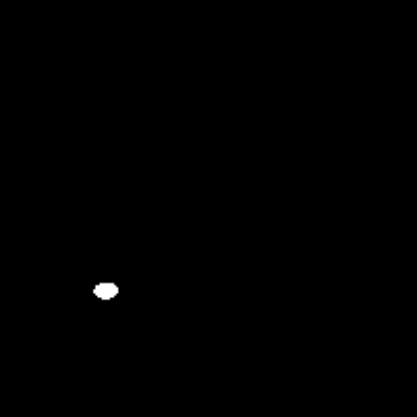} \\
\rotatebox{90}{Frame 200} & \includegraphics[width=0.25\textwidth]{car2GT200} & 
\includegraphics[width=0.25\textwidth]{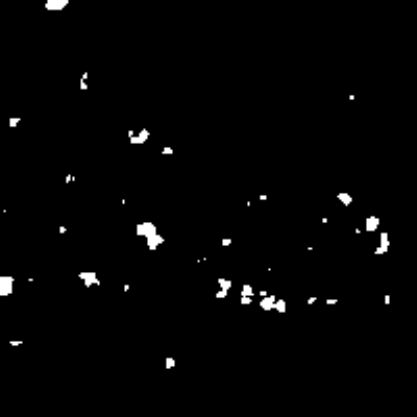} 
& \includegraphics[width=0.25\textwidth]{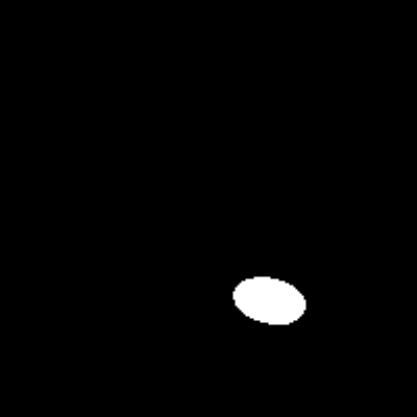}\\
\end{tabular}
\caption{Detection results on the Car2 sequence using the $TV-L^1$ optical flow. The left column is the groundtruth, the middle and right columns give the detected 
object from the original movement flow 
and geometric component of the flow, respectively.}
\label{fig:tvdetectcar2}
\end{figure}

\begin{figure}[!t]
\centering
\begin{tabular}{m{2mm}m{0.25\textwidth}m{0.25\textwidth}m{0.25\textwidth}}
\rotatebox{90}{Frame 20} & \includegraphics[width=0.25\textwidth]{car4GT20} & 
\includegraphics[width=0.25\textwidth]{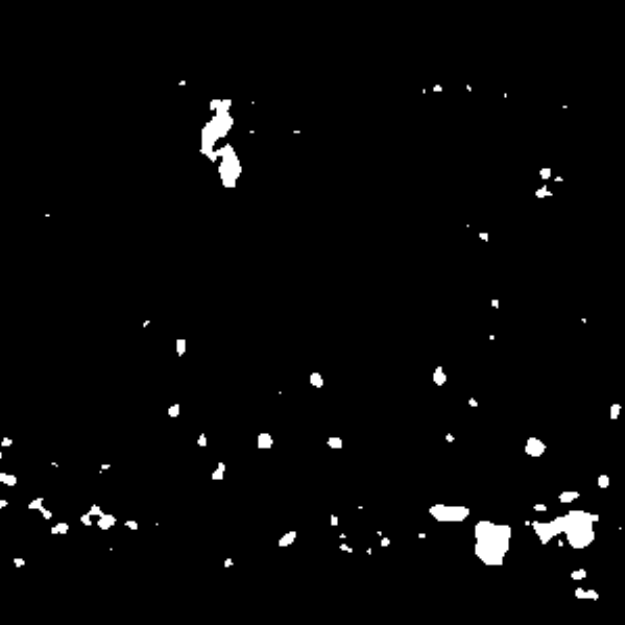} & \includegraphics[width=0.25\textwidth]{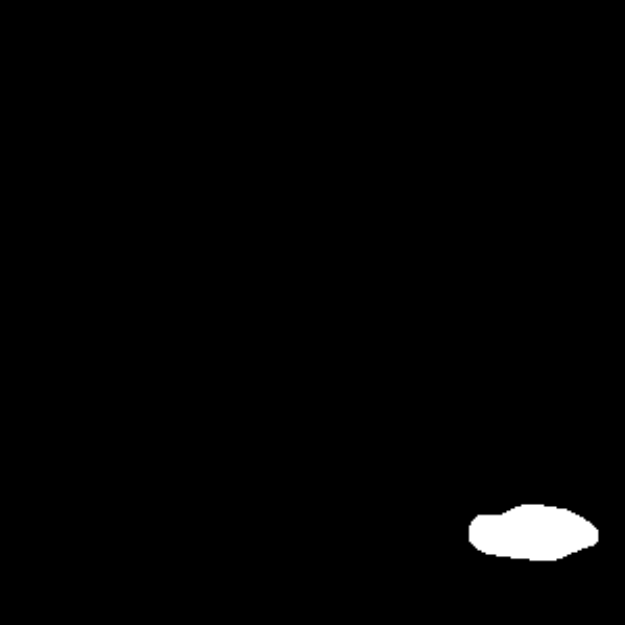} \\
\rotatebox{90}{Frame 40} & \includegraphics[width=0.25\textwidth]{car4GT40} & \includegraphics[width=0.25\textwidth]{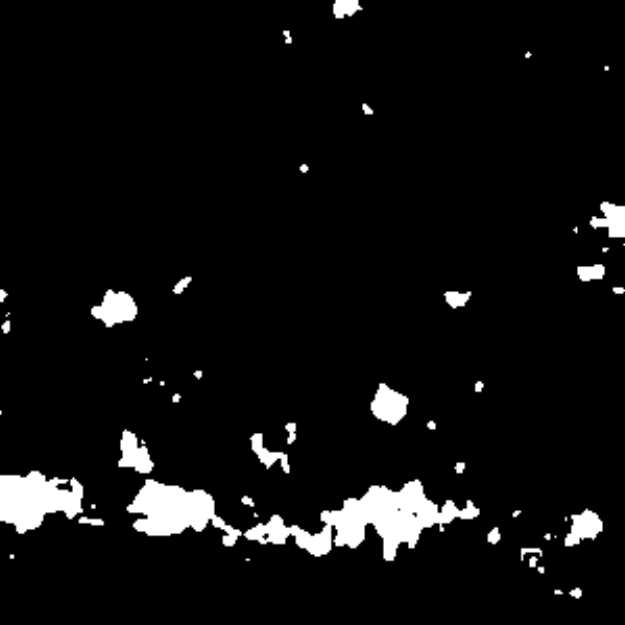} 
& \includegraphics[width=0.25\textwidth]{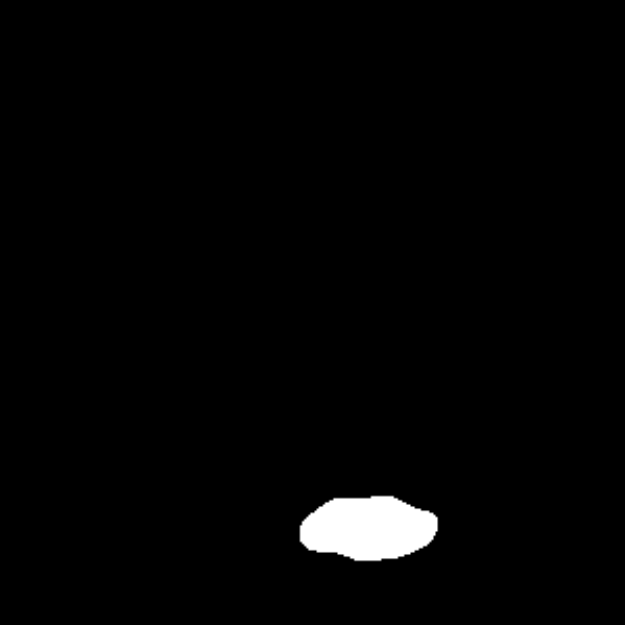}\\
\end{tabular}
\caption{Detection results on the Car4 sequence using the $TV-L^1$ optical flow. The left column is the groundtruth, the middle and right columns give the detected 
object from the original movement flow 
and geometric component of the flow, respectively.}
\label{fig:tvdetectcar4}
\end{figure}

In the previous sections, all experiments were performed by using the diffeomorphic Demons 
algorithm \cite{DiffeoDemons} to estimate the movement flows. In this section, we study the influence of the choice of the flow estimation technique. We propose to run the same 
experiments as in the previous section but by using two other movement flow estimation technique. We selected two optical-flow techniques: the classic multiscale Horn-Schunck 
approach \cite{Horn1981},\cite{ipolHS} and the more recent $TV-L^1$ model \cite{Zach},\cite{ipolTVL1}. We chose these techniques because they are popular techniques and have different characteristics. The 
Horn-Schunck approach is widely used in the image processing/computer vision community. It is based on the assumption of intensity consistency from one frame to the next one. The 
multiscale framework is used to allow the algorithm to take care of large movements. This model provides smooth flows and is sensitive to the presence of noise. The $TV-L^1$ model 
is a generalization of the Horn-Schunck model where the $TV$ regularization allows flows with discontinuity and the $L^1$ data fidelity term permits to drastically improve the 
stability against noise. The Demons algorithm corresponds to a completely different approach where the deformation flow is expected to be a diffeomorphism which implies a certain 
level of smoothness of the flow and captures very low magnitude flows. The results obtained by using the Demons algorithm were already given in the previous section and correspond to 
Figures~\ref{fig:detectcar1},\ref{fig:detectcar2} and \ref{fig:detectcar4}. The results obtained while using the Horn-Schunck or $TV-L^1$ optical flows are given in 
Figures~\ref{fig:hsdetectcar1},\ref{fig:hsdetectcar2} and \ref{fig:hsdetectcar4}, and Figures~\ref{fig:tvdetectcar1},\ref{fig:tvdetectcar2} and \ref{fig:tvdetectcar4}, 
respectively. We can observe that indeed the Horn-Schunck algorithm can be highly perturbed by the presence of turbulence and sometime is not even
capable of capturing the moving target at all. The $TV-L^1$ algorithm performs quite well and may have an advantage over the Demons algorithm because, the moving target corresponds to a 
piecewise constant flow, the $TV$ regularization enhances such characteristic. Moreover, with the $TV-L^1$ model, the turbulence seems to have less impact on the 
``randomness'' of the information in the flow. Even if the $TV-L^1$ looks to be better adapted for moving target detection through turbulence purposes, we can still observe some outliers in the original 
flow which are completely removed by our decomposition algorithm.



\subsection{Comparison with background subtraction technique}
\begin{figure}[!t]
\centering
\begin{tabular}{m{2mm}m{0.25\textwidth}m{0.25\textwidth}m{0.25\textwidth}}
\rotatebox{90}{Frame 10} & \includegraphics[width=0.25\textwidth]{car1GT10} & 
\includegraphics[width=0.25\textwidth]{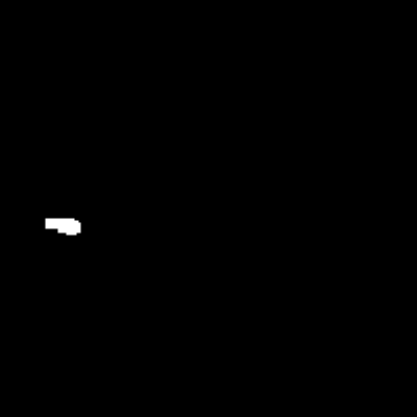} & \includegraphics[width=0.25\textwidth]{car1TVL1detectU10} \\
\rotatebox{90}{Frame 30} & \includegraphics[width=0.25\textwidth]{car1GT30} & \includegraphics[width=0.25\textwidth]{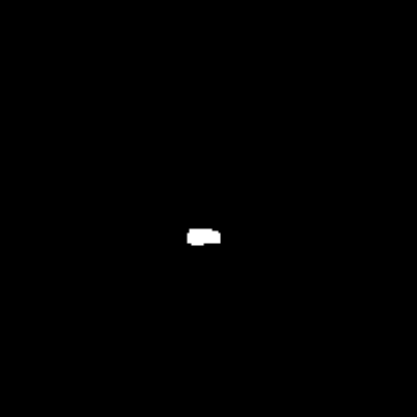} 
& \includegraphics[width=0.25\textwidth]{car1TVL1detectU30}\\
\end{tabular}
\caption{Comparison with Chen et al. background subtraction approach on the Car1 sequence. The left column is the groundtruth, the middle 
and right columns give Chen's results 
and our algorithm, respectively.}
\label{fig:chencar1}
\end{figure}

\begin{figure}[!t]
\centering
\begin{tabular}{m{2mm}m{0.25\textwidth}m{0.25\textwidth}m{0.25\textwidth}}
\rotatebox{90}{Frame 150} & \includegraphics[width=0.25\textwidth]{car2GT150} & 
\includegraphics[width=0.25\textwidth]{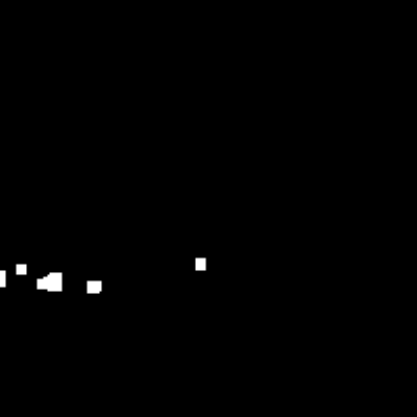} & \includegraphics[width=0.25\textwidth]{car2TVL1detectU150} \\
\rotatebox{90}{Frame 200} & \includegraphics[width=0.25\textwidth]{car2GT200} & 
\includegraphics[width=0.25\textwidth]{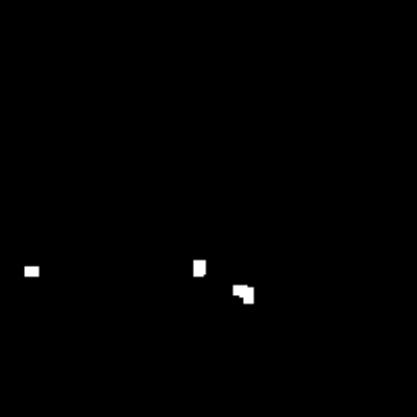} 
& \includegraphics[width=0.25\textwidth]{car2TVL1detectU200}\\
\end{tabular}
\caption{Comparison with Chen et al. background subtraction approach on the Car2 sequence. The left column is the groundtruth, the middle 
and right columns give Chen's results 
and our algorithm, respectively.}
\label{fig:chencar2}
\end{figure}

\begin{figure}[!t]
\centering
\begin{tabular}{m{2mm}m{0.25\textwidth}m{0.25\textwidth}m{0.25\textwidth}}
\rotatebox{90}{Frame 20} & \includegraphics[width=0.25\textwidth]{car4GT20} & 
\includegraphics[width=0.25\textwidth]{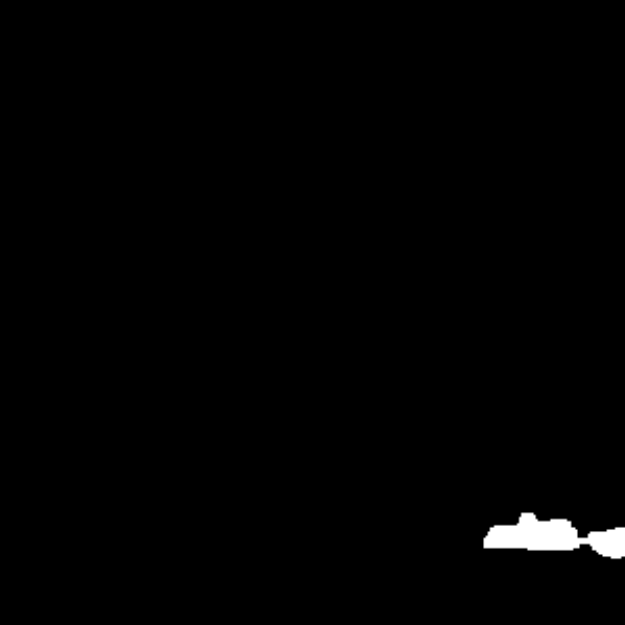} & \includegraphics[width=0.25\textwidth]{car4TVL1detectU20} \\
\rotatebox{90}{Frame 40} & \includegraphics[width=0.25\textwidth]{car4GT40} & \includegraphics[width=0.25\textwidth]{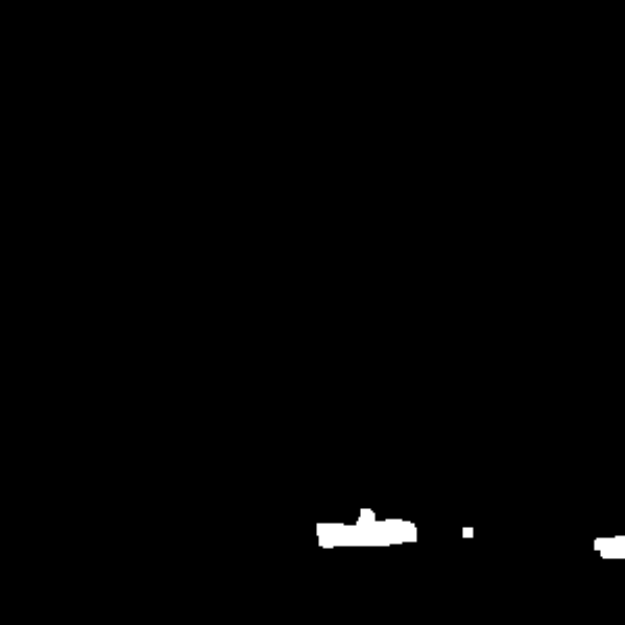} 
& \includegraphics[width=0.25\textwidth]{car4TVL1detectU40}\\
\end{tabular}
\caption{Comparison with Chen et al. background subtraction approach on the Car4 sequence. The left column is the groundtruth, the middle 
and right columns give Chen's results 
and our algorithm, respectively.}
\label{fig:chencar4}
\end{figure}

Due to their simplicity, background subtraction techniques are the most common methods used to detect moving objects. In \cite{Chen2014}, Chen et al. proposed to use this type of 
approach with an adaptive threshold in order to distinguish turbulence induced moving pixels and a moving target. Then combined with a specific tracking algorithm, they obtain 
satisfactory results. Since in this paper, we do not investigate the final tracking step, we only compare our detection results (using the $TV-L^1$ optical flow algorithm) with the background 
subtraction detection step as proposed in \cite{Chen2014}. The results are provided in Figures~\ref{fig:chencar1},\ref{fig:chencar2} and \ref{fig:chencar4}, respectively. We also 
invite the reader to watch the movies corresponding to the detections based on the original $TV-L^1$ flow 
(car1-ODetect.avi\footnote{\url{http://jegilles.sdsu.edu/movies/VFD/car1-ODetect.avi}}, car2-ODetect.avi\footnote{\url{http://jegilles.sdsu.edu/movies/VFD/car2-ODetect.avi}}, 
car4-ODetect.avi\footnote{\url{http://jegilles.sdsu.edu/movies/VFD/car4-ODetect.avi}}), the geometric component 
(car1-UDetect.avi\footnote{\url{http://jegilles.sdsu.edu/movies/VFD/car1-UDetect.avi}}, car2-UDetect.avi\footnote{\url{http://jegilles.sdsu.edu/movies/VFD/car2-UDetect.avi}}, 
car4-UDetect.avi\footnote{\url{http://jegilles.sdsu.edu/movies/VFD/car4-UDetect.avi}}) and Chen et al algorithm 
(car1-ChenDetect.avi\footnote{\url{http://jegilles.sdsu.edu/movies/VFD/car1-ChenDetect.avi}}, 
car2-ChenDetect.avi\footnote{\url{http://jegilles.sdsu.edu/movies/VFD/car2-ChenDetect.avi}}, 
car4-ChenDetect.avi\footnote{\url{http://jegilles.sdsu.edu/movies/VFD/car4-ChenDetect.avi}}). If Chen et al algorithm performs well in most cases, we can notice two major 
drawbacks. First, on the hardest sequence (Car2), many outliers are still present in the image. Second, sometimes the moving object is not detected at all, this 
could be a serious problem for a tracking algorithm which comes next since that ``breaks'' the trajectory continuity. Our method clearly outperforms Chen et al algorithm as it 
detects continuous trajectory without any outliers.



\section{Conclusions}\label{sec:conc}
In this paper, we addressed the detection of moving objects when observations are impacted by the atmospheric turbulence. We presented a new geometric point of view where 
consistent movement of objects corresponds to piecewise constant patterns in the 3D spatio-temporal velocity field while turbulence has an oscillating behavior. We thus extended 
existing decomposition methods to process vector fields. We want to emphasize that if a simple temporal filtering could be able to extract part of the expected target, our 
approach will give better results since the regularization applies with respect to both space and time. Moreover, a linear filtering will necessarily blur the area 
corresponding to a potential target while the geometric component provided by our approach keep sharp edges. The experiments show that the proposed algorithm is capable of 
separating the two kinds of movement to clearly improve 
the detection results which will make the tracking easier to perform. The proposed algorithm could be easily generalized to any type of vector fields and in any dimension (up to the code availability to perform the wavelet/curvelet transform in 
higher dimensions). Other potential applications could be in the analysis of experimentally acquired fluid velocities in order to characterize mechanical properties. In terms of 
future work, we will investigate the use of the recently-developed empirical wavelet transform \cite{EWT1D,EWT2D} that permits to have an adaptive curvelet representation. 
Unfortunately, the 3D version of that transform is not yet available and we are currently developing the code for the 3D version of this transform.

\section*{Acknowledgement}
The authors want to thank the anonymous reviewers for their comments and suggestions which permit to deeply improve the manuscript. This material is based upon work supported by 
the Air Force Office of Scientific Research under award number FA9550-15-1-0065 and the NSF grant DMS-1556480.

\bibliography{references}

\end{document}